\documentclass{article} % For LaTeX2e
\usepackage{iclr2023_conference,times}

\usepackage{amsmath,amsfonts,bm}

\def\eqref#1{equation~\ref{#1}}
\def\1{\bm{1}}

\DeclareMathAlphabet{\mathsfit}{\encodingdefault}{\sfdefault}{m}{sl}
\SetMathAlphabet{\mathsfit}{bold}{\encodingdefault}{\sfdefault}{bx}{n}

\usepackage{hyperref}
\usepackage{url}

\newcommand{\xjqi}[1]{{\color{orange}{\bf\sf [xjqi: #1]}}}
\newcommand{\lzz}[1]{{\color{purple}{\bf\sf [lzz: #1]}}}
\newcommand{\phil}[1]{{\color{blue}{#1}}}

\newcommand{\ie}{\textit{i.e.}}
\newcommand{\eg}{\textit{e.g.}}

\usepackage[utf8]{inputenc} % allow utf-8 input
\usepackage[T1]{fontenc}    % use 8-bit T1 fonts
\usepackage{hyperref}       % hyperlinks
\usepackage{url}            % simple URL typesetting
\usepackage{booktabs}       % professional-quality tables
\usepackage{amsfonts}       % blackboard math symbols
\usepackage{nicefrac}       % compact symbols for 1/2, etc.
\usepackage{microtype}      % microtypography
\usepackage{xcolor}         % colors
\usepackage[T1]{fontenc}
\usepackage[linesnumbered,boxed,ruled,commentsnumbered]{algorithm2e}
\usepackage{times}
\usepackage{stfloats}
\usepackage{epsfig}
\usepackage{graphicx}
\usepackage{amsmath}
\usepackage{amssymb}
\usepackage[utf8]{inputenc} % allow utf-8 input
\usepackage[T1]{fontenc}    % use 8-bit T1 fonts
\usepackage{url}            % simple URL typesetting
\usepackage{booktabs}       % professional-quality tables
\usepackage{amsfonts}       % blackboard math symbols
\usepackage{nicefrac}       % compact symbols for 1/2, etc.
\usepackage{microtype}      % microtypography
\usepackage{graphicx}
\usepackage{multirow}
\usepackage{wrapfig}
\usepackage{xcolor}
\usepackage{bbm}
\usepackage{amsmath}
\usepackage{comment}
\usepackage{dsfont}
\usepackage{bbm}
\usepackage{wrapfig}
\usepackage{float}
\usepackage{enumitem}
\usepackage{cancel}
\usepackage{lipsum}

\usepackage{hyperref}
\let\svthefootnote\thefootnote
\newcommand\freefootnote[1]{%
  \let\thefootnote\relax%
  \footnotetext{#1}%
  \let\thefootnote\svthefootnote%
}

\title{ISS: Image as Stepping Stone\\ for Text-Guided 3D Shape Generation}

\author{Zhengzhe Liu$^{1}$ \quad   Peng Dai$^{2}$  \quad  Ruihui Li$^{3}$ \quad    Xiaojuan Qi$^{2*}$   \quad   Chi-Wing Fu$^{1*}$ \\
$^1$The Chinese University of Hong Kong \quad $^2$The University of Hong Kong \quad $^3$Hunan University \\
}

\iclrfinalcopy % Uncomment for camera-ready version, but NOT for submission.
\begin{document}
%\blfootnote{*: Corresponding authors}
%\freefootnote{*: Corresponding authors.}
\maketitle
\begin{abstract}

Text-guided 3D shape generation 
remains challenging due to the absence of large paired text-shape dataset, the substantial semantic gap between these two modalities, and the structural complexity of 3D shapes.
This paper presents a new framework called {\em Image as Stepping Stone\/} (ISS) for the task by introducing 2D image as a stepping stone to connect the two modalities and to eliminate the need for paired text-shape data.
Our key contribution is a {\em two-stage feature-space-alignment approach\/} that maps CLIP features to shapes by harnessing a pre-trained single-view reconstruction (SVR) model with multi-view supervisions: first map the CLIP image feature to the detail-rich shape space in the SVR model, then map the CLIP text feature to the shape space 
%with a fast test-time optimization
and optimize the mapping by encouraging 
%that encourages 
CLIP consistency between the input text and the rendered images.
%between the rendered images and the input text.
Further, we formulate a {\em text-guided shape stylization module\/} to dress up the output shapes with novel structures and textures. 
%we design a text-guided shape stylization module to generate novel textures on the output shapes.
Beyond existing works on 3D shape generation from text, our new approach is general for creating shapes in a broad range of categories, {\em without\/} requiring paired text-shape data.
%Also, it can work  with different single-view reconstruction models.
%
%\textcolor{purple}{Beyond a specific solution to text-to-shape generation, our approach is a general solution to effectively reduce this challenging task to an easier task single view reconstruction (SVR), which has been more intensively studied.}
%\phil{Is this sentence very important and exciting in a high-level sense? Remove it?}
%
%Experimental results manifest that our approach outperforms the state-of-the-arts and
%existing works and 
%our baselines in terms of {\em fidelity\/}, {\em diversity\/}, 
%{\em generative novelty\/}, and {\em consistency between text and shape\/}.
%\phil{any exciting and strong thing that we may say about our results?}
%\lzz{We may not surpass Dream Field in terms of diversity and generative novelty (Actually it is easier to generate multi-view images with better diversity and generative novelty than 3D shapes). In our experiment, we only measure fidelity (FID, FPD) and consistency
%(the user study) quantitatively. May say "generative efficiency (speed)", "realism", "functionality", "stylization"? }
%\phil{I see. Then, we should not overclaim}
%
Experimental results manifest that our approach outperforms the state-of-the-arts and
our baselines in terms of {\em fidelity\/} and {\em consistency with text}. Further, our approach can stylize the generated shapes with both realistic and fantasy structures and textures.
Codes are available at 
{\scriptsize \url{https://github.com/liuzhengzhe/ISS-Image-as-Stepping-Stone-for-Text-Guided-3D-Shape-Generation}}.
%\phil{Is it good to highlight the speed of our approach?}\lzz{It is ok to say "with a fast test-time optimization", but may not highlight the speed, since we need test-time optimization, while CLIP-Forge not need. }\phil{I see}
%
%\lzz{Code release is mentioned in Introduction, may not need to mention it here? }
\end{abstract}

\begin{figure}[H]
\centering
\includegraphics[width=0.99\textwidth]{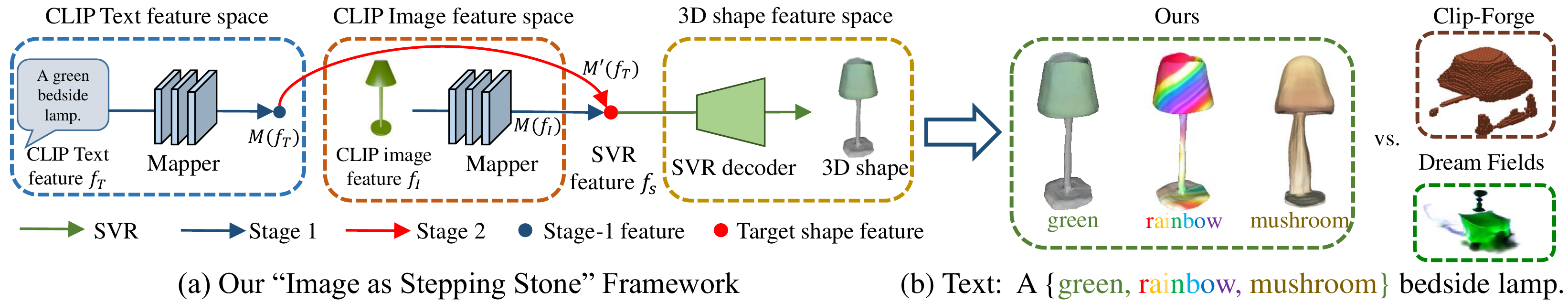}
\vspace*{-2mm}
\caption{Our novel ``Image as Stepping Stone'' framework 
%two-stage feature-space-alignment approach 
(a) is able to connect the text space (the CLIP Text feature) and the 3D shape space (the SVR feature) %generate 3D shapes from texts 
through our two-stage feature-space alignment, 
%with 2D image as a stepping stone, 
such that we can generate plausible 3D shapes from text (b) beyond the capabilities of the existing works (CLIP-Forge and Dream Fields), without requiring paired text-shape data.
}
\label{fig:figure1}
\vspace*{-2.75mm}
\end{figure}

%\vspace*{-5pt}
\section{Introduction}\label{sec:intro}

3D shape generation has a broad range of applications,~\eg, in Metaverse, CAD, games, animations, etc.
Among various ways to generate 3D shapes, a user-friendly approach is to generate shapes from natural language or text descriptions.
%it-you-say-is-what-you-get fashion. 
By this means, users can readily create shapes,~\eg, to add/modify objects in VR/AR worlds, to design shapes for 3D printing, etc. %\xjqi{better have some examples}. 
%
%Although considerable progress \phil{is it alright to say ``considerable progress'' cause there are only a few works, right?  If not, start this paragraph with something else, e.g., ``However, it is nontrivial to generate 3D shapes from text descriptions'' or ``However, generating shapes from text descriptions is a very challenging problem.  Existing works ...''}
%
Yet, generating shapes from texts is very challenging, due to the lack of large-scale paired text-shape data, the large semantic gap between the text and shape modalities, and the structural complexity 
%irregularity 
%\phil{complexity?}
of 3D shapes.

Existing works~\citep{chen2018text2shape,jahan2021semantics,liu2022towards} typically rely on paired text-shape data for model training. Yet, collecting 3D shapes is already very challenging on its own, let alone the tedious manual annotations needed to construct the text-shape pairs.  To our best knowledge, the largest existing paired text-shape dataset~\citep{chen2018text2shape} contains only two categories,~\ie, table and chair, thus severely limiting the applicability of the existing works.

Very recently, two annotation-free approaches, CLIP-Forge~\citep{sanghi2021clip} and Dream Fields~\citep{jain2021zero}, were proposed to address the dataset limitation.
These two state-of-the-art approaches attempt to
%They attempt to %\phil{attempt to? cause they didn't completely solve the problem, right?} 
utilize the joint text-image embedding from the large-scale pre-trained language vision model,~\ie, CLIP~\citep{radford2021learning}, to eliminate the need of requiring paired text-shape data in model training. However, it is still extremely challenging to generate 3D shapes from text without paired texts and shapes for the following reasons. First, the range of object categories that can be generated are still limited due to the scarcity of 3D datasets. For example, Clip-Forge~\citep{sanghi2021clip} is built upon a shape auto-encoder;
%, so it still relies on shapes for training
it can hardly generate 
%reasonable 
plausible shapes beyond the ShapeNet categories. 
Also, it is challenging to learn 3D prior of the desired shape from texts. For instance, Dream Field~\citep{jain2021zero} cannot generate 3D shapes like our approach due to the lack of 3D prior, as it is trained to produce only multi-view images with a neural radiance field. %even though leveraging the large-scale pre-trained language vision model,~\ie, CLIP~\citep{radford2021learning}.
Further, with over an hour of optimization for each shape instance from scratch, there is still no guarantee that the multi-view consistency constraint of Dream Field~\citep{jain2021zero} can enforce the model for producing shapes
%\textcolor{purple}{\cancel{3D shape}} \phil{this sentence is strange in the sense that you just said that DF is trained to produce only multi-view images... why it can generate 3D shapes?}
that match the given text; we will provide further investigation in our experiments.  
Last, the visual quality of the generated shapes is far from satisfactory due to the substantial semantic gap between the unpaired texts and shapes. As shown in Figure~\ref{fig:figure1} (b), the results generated by Dream Field typically look surrealistic (rather than real),
%are often far from realistic, 
%\phil{I tried not to say something too bad for DF}
%
due to insufficient information extracted from text for the shape structures and details.
%\phil{try to explain more on why no realistics... structurally, right? also?}
On the other hand, CLIP-Forge~\citep{sanghi2021clip} is highly restricted by the limited $64^3$ resolution and it lacks colors and textures, further manifesting the difficulty of generating 3D shapes from unpaired text-shape data.

Going beyond the existing works,
%In this work, 
we propose a 
%\textcolor{purple}{lightweight but powerful} 
%\phil{doesn't feel good to mention lightweight too early... perhaps later in the paper?}
%
new approach to 3D shape generation from text without needing paired text-shape data.
%\textcolor{purple}{going beyond existing works mentioned above}. 
%\phil{add one sentence here to emphasize that even it is lightweight, it has some advantages beyond existing works? Or mention certain strong results that ISS can achieve?}
%
Our key idea is to {\em implicitly leverage 2D image\/} as a stepping stone (ISS) to connect the text and shape modalities.
Specifically, we employ the joint text-image embedding in CLIP and train a CLIP2Shape mapper to {\em map the CLIP image features to a pre-trained detail-rich 3D shape space with multi-view supervisions\/}; see Figure~\ref{fig:figure1} (a): stage 1.
%
%\phil{not just refer to stage 1 but refer to stage 2 somewhere in this paragraph as well?} \lzz{see end of this paragraph. }
Thanks to the joint text-image embedding from CLIP, 
our trained mapper is able to connect
%we adopt the above mapper to %\phil{also? further?} 
%further map 
the CLIP text features with the shape space for text-guided 3D shape generation.
Yet, due to the gap between the CLIP text and CLIP image features, the mapped text feature may not align well with
%has a certain distance to 
the destination shape feature; see the empirical analysis in Section~\ref{sec:empirical}. %\phil{(see an empirical analysis in Section XXX)}.
%0.59 in Figure~\ref{fig:figure1} (d): stage 2).
Hence, we further fine-tune the mapper specific to each text input by {\em encouraging CLIP consistency\/} between the rendered images and the input text to enhance the consistency between the input text and the generated shape; see Figure~\ref{fig:figure1} (a): stage 2.
%\phil{refer Figure 1 (b) here?}
%\lzz{May refer Figure 1 (a): stage 2 here. }
%\phil{ok}

Our new approach %addresses
advances the frontier of 3D shape generation from text 
%goes beyond the two existing works
%the issues of existing works~\citep{sanghi2021clip,jain2021zero} 
in the following aspects.
%
%\phil{mention image as a stepping stone somehow in the sentence below?}
First, by taking image as a stepping stone, we make the challenging text-guided 3D shape generation task more approachable and cast it as a single-view reconstruction (SVR) task. 
%\textcolor{purple}{with image as stepping stone}, and 
Having said that, we learn 3D shape priors from the adopted SVR model directly in the feature space.
%, advancing the frontier of the 3D shape generative capability from texts.
%First, we leverage the joint text-image embeddings from CLIP and learn 3D shape priors from a pre-trained single-view reconstruction (SVR) model directly in the feature space, eliminating the need for the paired text-shape-data like~\citep{chen2018text2shape,liu2022towards}.
Second, benefiting from the learned 3D priors from the SVR model and the joint text-image embeddings, our approach can produce 3D shapes in only 85 seconds vs. 72 minutes of Dream Fields~\citep{jain2021zero}.
%, which can only produce multi-view images at the end.
%\phil{maybe don't say bad words about DF that much... they may be our reviewers}
%
More importantly, %with the two-stage feature space alignment that gradually reduces the gap among text, image and shape modalities, 
our approach is able to produce 
%high-quality 
plausible 3D shapes, {\em not\/} multi-view images, beyond the generation capabilities 
of the state-of-the-art approaches;
%CLIP-Forge and Dream Fields; 
see Figure~\ref{fig:figure1} (b).

With our two-stage feature-space alignment, we already can generate shapes with good fidelity from texts.
%as shown in Figure~\ref{fig:figure1} (c).
To further enrich the generated shapes with vivid textures and structures %\phil{enrich the generated shapes with vivid textures}
beyond the generative space of the pre-trained %\phil{pre-trained? or pre-trained? please standardize it in the whole paper} 
SVR model,
we additionally design a text-guided stylization module to generate novel textures and shapes by encouraging consistency between the rendered images and the text description of the target style. 
%\xjqi{briely summarize how can you do this}. 
We then can 
%seamlessly 
effectively fuse with the two-stage feature-space alignment to enable the generation of both realistic and fantasy textures and also shapes beyond %the training data 
the generation capability of the SVR model; see Figure~\ref{fig:figure1} (b) for examples.
Furthermore, our approach is compatible with different SVR models~\citep{niemeyer2020differentiable,alwala2022pre}.
For example, 
%For example, by adopting SS3D~\citep{alwala2022pre}, which 
we can adopt SS3D~\citep{alwala2022pre} to generate shapes from single-view in-the-wild images to broaden the range of categorical 3D shapes that our approach can generate, 
going beyond~\citet{sanghi2021clip}, which can only generate 13 categories of ShapeNet. Besides, our approach can also work with the very recent approach GET3D~\citep{gao2022get3d} to generate
high-quality 3D shapes from text; 
see our results in Section~\ref{sec:results}.
\section{Related Works}

% Philip: Usually, we put -3pt before paragraph... I did that in most of my papers before

%The remarkable success of single view reconstruction and the strong correlation between 2D image and 3D shape motivate us to reduce the challenging text-to-shape generation into single view reconstruction. Building upon a single view reconstruction model, our approach can effectively generate high-fidelity shapes from the text input. In this work, we build our approach on DVR~\citep{niemeyer2020differentiable} and SS3D~\citep{alwala2022pre} as representatives \phil{this work is not clear} for text-guided shape generation with multi-view and single-view images in training.

%\phil{(1) SVR is certainly related to this paper BUT SVR is kind of a image-guided 3D generation.  It has a clear prior, which is an image, for generating 3D shapes.  We need to explicitly say how SVR is related to this work and WHY it is different, cause some SVR papers can create pretty nice results compared to ours, since the input images are very good prior and provide very rich information for 3D shape generation; (2) }

\vspace*{-1mm}
\paragraph{Text-guided image generation.}
Existing text-guided image generation approaches can be roughly cast into two branches: (i) direct image synthesis~\citep{reed2016generative,reed2016learning,zhang2017stackgan,zhang2018stackgan++,xu2018attngan,li2019controllable,li2020manigan,qiao2019mirrorgan,wang2021cycle} and (ii) image generation with a pre-trained GAN~\citep{stap2020conditional,yuan2019bridge,souza2020efficient,wang2020text,rombach2020network,patashnik2021styleclip,xia2021tedigan}. Yet, the above works can only generate images for limited categories. To address this issue, some recent works explore zero-shot text-guided image generation~\citep{ramesh2021zero,ding2021cogview,nichol2021glide,liu2021fusedream,ramesh2022hierarchical} to learn to produce images of any category. Recently,~\citet{zhou2021lafite} and ~\citet{wang2022clip} leverage CLIP for text-free text-to-image generation.
%\phil{TODO: (1) please fix some ? above and (2) it is strange when reading this paragraph word by word, as some author names are duplicated: Zhou et al. followed by Zhou et al.; I guess you may simply say~\citep{zhou2021lafite} rather than Zhou~\etal~\citep{zhou2021lafite}}
Text-guided shape generation is more challenging compared with text-to-image generation. First, it is 
%extremely hard and labor-intensive 
far more labor-intensive and difficult
to prepare a large amount of paired text-shape data than paired text-image data, which can be collected from the Internet on a large scale.
Second, the text-to-shape task requires one to predict full 3D structures that %\phil{should be}
should be plausible geometrically and 
consistently in all views, %\phil{requires one to predict full 3D structures that are plausible consistently in all views},
%360-degree prediction of shapes with structural consistency in 3D,
beyond the needs in single-view image generation.
Third, 3D shapes may exhibit more complex spatial structures and topology, beyond regular grid-based 2D images.

\vspace*{-6mm}
\paragraph{Text-guided 3D generation.}
To generate shapes from text, several works~\citep{chen2018text2shape,jahan2021semantics,liu2022towards} rely on paired text-shape data for training.
To avoid paired text-shape data, 
%Clearly, collecting 3D shapes is extremely laborious and challenging, let alone the manual works to annotate them with natural language descriptions.
two very recent works, CLIP-Forge~\citep{sanghi2021clip} and Dream Fields~\citep{jain2021zero}, attempt to leverage the large-scale pre-trained vision-language model CLIP.
%eliminate the requirement of paired text-shape data.
%models. 
Yet, they still suffer from various limitations, as discussed in the third paragraph of Section~\ref{sec:intro}. 
Besides 3D shape generation, some recent works utilize CLIP to manipulate a shape or NeRF with text~\citep{michel2021text2mesh,jetchev2021clipmatrix,wang2021clipnerf} and to generate 3D avatars~\citep{hong2022avatarclip}.
In this work, we present a new framework for generating 3D shape from text without paired text-shape data by our novel two-stage feature-space alignment.
%\phil{mentions the techniques a bit}
Our experimental results demonstrate the superiority of this work beyond the existing ones in terms of fidelity and text-shape consistency.
%\phil{diversity? generative novelty?} \lzz{We may not surpass Dream Field in terms of diversity and generative novelty. In our experiment, we only measure fidelity (FID, FPD) and consistency (the
%user study) quantitatively. May say "realism", "functionality", "stylization"?}

\vspace*{-6mm}
\paragraph{Single-view reconstruction.} %Image-to-Shape Generation}
%\phil{(1) Move this SVR subsection to the end of Section 2?   It is because this subsection is the least related compared with the other two and we also don't want the primary to find reviewers in this area.
%(2) Try to shorten this subsection (esp. the second paragraph here).
%(3) start it like this: Another topic that is related to this work is single-view reconstruction, in which researchers have explored SVR with meshes...}
%
Another topic related to this work is single-view reconstruction (SVR).
Recently, researchers have explored SVR with meshes~\citep{agarwal2020gamesh}, voxels~\citep{zubic2021effective}, and 3D shapes~\citep{niemeyer2020differentiable}. 
%Recent years have witnessed the remarkable progress of single-view reconstruction (SVR) for meshes~\citep{wang2018pixel2mesh,chen2019learningto3d,chen2020bsp,agarwal2020gamesh}, voxels~\citep{xie2019pix2vox,zubic2021effective}, and implicit 3D shapes~\citep{mescheder2019occupancy,chen2019learning,xu2019disn,jiang2020sdfdiff,li2020d,niemeyer2020differentiable}. %Most of the above works utilize multi-view images for training.
Further, to extend SVR to in-the-wild categories,~\citet{alwala2022pre} propose SS3D to learn 3D shape reconstruction using single-view images in hundreds of categories. 
In our work, we propose to harness an SVR model to map images to shapes, such that we can take 2D image as a stepping stone for producing shapes from texts.
Yet, we perform the mapping and feature alignment implicitly in the latent space rather than explicitly.

%\textcolor{purple}{As 3D-to-2D projections, single-view 2D images are more closely related to shapes than texts, since they reveal many attributes of 3D shapes,~\eg, structure, details, appearance, etc.
%The strong correlation between 2D images and 3D shapes motivates us to reduce the challenging text-to-shape generation task to 
%text-to-image then SVR, by connecting the CLIP features from text with the shape features in SVR.}
%Specifically, 
%our framework can work with different SVR approaches to extend them for 3D shape generation from texts.
%So, our approach is orthogonal 
%to the SVR approaches.
%Specifically, we build our approach on DVR~\citep{niemeyer2020differentiable} and SS3D~\citep{alwala2022pre} as two representatives, which take multi- and single-view images in training, respectively.

%however, the generated fidelity are far from satisfying, and the need of 3D shapes for training restricts their generation to a handful of categories. Another recent work Dream Fields~\citep{jain2021zero} handles shape generation in a wide range of categories from text. However, it only train a NeRF, and cannot produce concrete 3D shapes directly. The unsatisfying generation speed and unlifelike generative quality make it hard to be practical utilized. 

%\phil{please double check and ensure all recent related works have been citepd}

%%%%%%%%%%%%%%%%%%%%%%%%%%%%%%%%%%%%%%%%%%%%%%%%%%%%%%%%%%%%%%%%%%%%%%
\vspace*{-5pt}
\section{Methodology}

%\vspace*{-3pt}
\begin{figure*}
\centering
\includegraphics[width=0.99\textwidth]{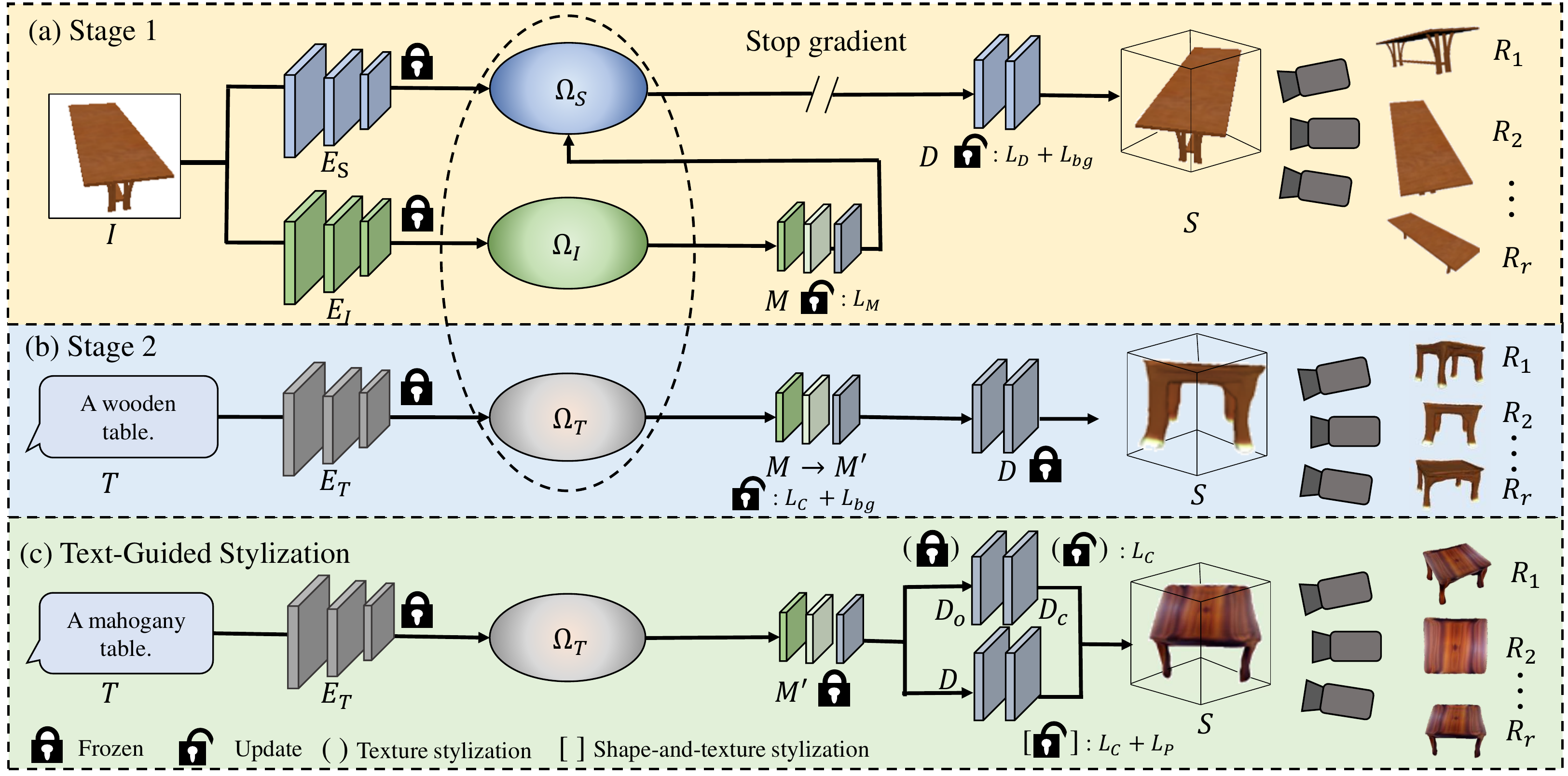}
\vspace*{-2mm}
\caption{%\xjqi{revise the figure to make it compact}
Overview of our text-guided 3D shape generation framework, which has three major 
%parts.
stages.
(a) Leveraging a pre-trained SVR model, in stage-1 feature-space alignment, we train the CLIP2Shape mapper $M$ to map the CLIP image feature space $\Omega_{\text{I}}$ 
%to the shape space \phil{of the SVR model, obtain} $\Omega_{\text{S}}$,
to shape space $\Omega_{\text{S}}$
%$\Omega_{\text{S}}$ 
of the SVR model with $E_\text{S}$, $E_\text{I}$ frozen,
%\phil{$\Omega_{\text{S}}$ looks like features with F as a symbol like $\Omega_{\text{I}}$; use some other symbol for space?}\lzz{how about $\Omega$?}
and fine-tune decoder $D$ with an additional background loss $L_{\text{bg}}$. $M$ and $D$ are trained with their own losses separately at the same time by stopping the gradients from SVR loss $L_D$ and background loss $L_{bg}$ to propagate to $M$.
(b) In stage-2 feature-space alignment, we fix $D$ and %\phil{freeze $D$ and} 
fine-tune $M$ into $M'$ by encouraging CLIP consistency between input text $T$ and the rendered images at test time.
(c) Last, we optimize the style of the 
%optimally stylize the 
generated shape and texture of $S$ for $T$. %\phil{for $T$?}
%text input $T$.
At the inference, we use stage 2 to generate 3D shape from $T$ and (c) is optional.
%\phil{please update the symbols in figure caption: $F$ to $\Omega$}
%\pdai{Should the shape of the table in s and (c) remain the same?}
%\phil{please move every symbol and text label closer to the image that it describes,~\eg, $E_I$ is quite far from its associated image;
%also center align every text label with its associated image}
%\phil{I now understand that you keep using $F_{something}$ to mean feature space... that's okay but perhaps $\Omega$ is better than $F$, you may decide but need to make it consistent everywhere}
}

\label{fig:overview}
\vspace*{-2.5mm}
\end{figure*}

\subsection{Overview}

This work aims to generate 3D shape $S$ from text $T$. 
%\phil{Overall, our idea is to connect and map the CLIP image features to the shape space of a pre-trained SVR model.}
Overall, our idea is to map the CLIP features to the shape space of a pre-trained SVR model, such that we can leverage the joint text-image embeddings from CLIP and also the 3D generation capability of the SVR model to enhance the generation of 3D shape from text.
Hence, our method only needs to be trained with multi-view RGB or RGBD images and the associated camera poses without paired %\phil{better to add ``paired'' here? for consistently with what we have said so far in the paper} 
text-shape data. %to train the CLIP2Shape mapper $M$ \lzz{composed of 12 fully-connected layers, each followed by a Leaky-ReLU}. %\phil{to train the 
%in the training, we only need multi-view images with associated camera poses without text and shape to train the CLIP2Shape mapper $M$ \lzz{composed of 12 fully-connected layers, each followed by a Leaky-ReLU}. %\phil{to train the CLIP2Shape mapper $M$}.
As Figure~\ref{fig:overview} shows, our framework 
%based upon a pre-trained SVR frameworks which has 
includes (i) image encoder $E_{\text{S}}$, which maps input image $I$ to SVR shape space $\Omega_{\text{S}}$, 
(ii) pre-trained CLIP text and image encoders $E_{\text{T}}$ and $E_{\text{I}}$, which map text $T$ and image $I$ to CLIP spaces $\Omega_{\text{T}}$ and $\Omega_{\text{I}}$, respectively,
(iii) mapper $M$ with 12 fully-connected layers, each followed by a Leaky-ReLU,
and (iv) decoder $D$ to generate the final shape $S$.
%\phil{mention what $E_I$? Why only need to mention $E_S$ here?}
Specifically, we use DVR~\citep{niemeyer2020differentiable} as the SVR model when presenting our method, unless otherwise specified.
%if not specified. 
%\phil{NOTE: I think pre-trained instead of pre-trained; please fix this throughout the paper.
%You may try to google ``pre-trained'' and google ``pre-trained'' and see the counts}

%First of all, we finetune $D$ with the background loss as shown in Figure~\ref{fig:overview} (a). T
%In more detail, 
%\phil{In short,} 
Overall, we introduce a novel two-stage feature-space-alignment approach to connect the text, image, and shape modalities. In detail, we first train CLIP2Shape 
%
%(C2S) \phil{remove this? looks too much cause CLIP2Shape is clear}
%
mapper $M$ to connect the CLIP image space $\Omega_\text{I}$ and the shape space $\Omega_\text{S}$ from the pre-trained SVR model %\phil{from the pre-trained SVR model} 
(see Figure~\ref{fig:overview} (a)).
Then, we fine-tune $M$ at test time using a CLIP consistency loss $L_c$ to further connect the CLIP text space $\Omega_\text{T}$ with $\Omega_\text{S}$ (see Figure~\ref{fig:overview} (b)). Last, we may further %\phil{may further} 
optimize the texture and structure
style of $S$ by fine-tuning the decoders (see Figure~\ref{fig:overview} (c)). %color decoder $D_c$ with occupancy decoder $D_o$ frozen (see Figure~\ref{fig:overview} (c)).

In the following, we first introduce two empirical studies on the CLIP feature space %\phil{NOTE: no hyphen, right?  cause it is a noun, not adj.} 
in Section~\ref{sec:empirical}, then present our two-stage feature-space-alignment approach in Section~\ref{sec:two-sgage}. Further, Section~\ref{sec:stylization} presents our text-guided shape stylization method and Section~\ref{sec:compatible} discusses the compatibility of our approach with different SVR models %\phil{with ... compatible with what?} 
and our %\phil{our?}
extension to generate a broad range of categories. 

\vspace*{-3pt}
\subsection{Empirical Studies and Motivations}\label{sec:empirical}

%\xjqi{please modify this section as discussed. Besides, it's better to move numbers to a table or use visualization to make it easier to understand}

Existing works~\citep{sanghi2021clip,zhou2021lafite,wang2022clip} mostly %\phil{mostly?} 
utilize CLIP directly %\phil{directly} 
%in their applications
without analyzing 
%but not analyze 
how it works and discussing its limitations. To start, we investigate the merits and drawbacks of leveraging CLIP for text-guided 3D shape generation by conducting the following  %\phil{the following} 
two empirical studies to gain more insight into the CLIP feature space. 
%a better understanding of the CLIP feature space. 

%please modify this section as discussed. 
%\phil{who wrote this? revised?}\lzz{xiaojuan}
%\phil{revised? Is this subsection ready for me to work on?}\lzz{yes. She has finished revision.}
%\phil{Got it!}

\begin{comment}
\begin{wrapfigure}{r}{0.3\textwidth}
  \begin{center}
    \includegraphics[width=0.28\textwidth]{motivation.pdf}
  \end{center}
  \caption{Single image reconstruction by DVR and $E_{\text{I}}$+$D$. }
  \label{fig:motivation}
\end{wrapfigure}
\end{comment}

\begin{figure*}
\centering
\includegraphics[width=0.99\textwidth]{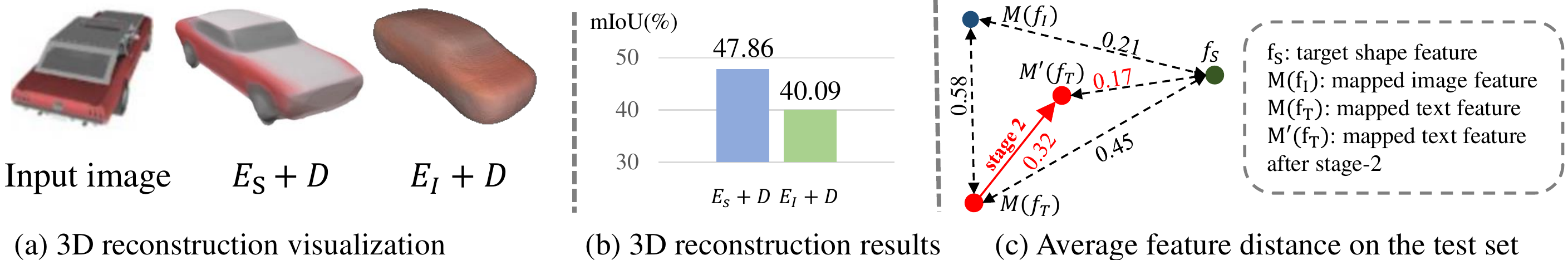}
\vspace*{-1.5mm}
\caption{Empirical studies on the CLIP feature space for text-guided 3D shape generation. %\pdai{How to obtain the numbers in (c)?}
%\phil{In (a): standardize the gaps around the two plus signs;
%(b): just photoshop the image such that the blue and green boxes and centered with the text labels above and below each of them?
%(c) stage-2 -> stage 2;}
%what is the distance from $M(f_T)$ to $f_s$?  If you need to add a number for this link, move 0.59 below its associated arrow}
} %\phil{on CLIP feature space for text-guided 3D shape generation}.}
%\phil{write more to describe what's going on above?}}
\label{fig:motivation}
\vspace*{-2.5mm}
\end{figure*}

\vspace*{-3pt}
\subsubsection{
Whether % pre-trained
CLIP feature space sufficiently good for 3D shape generation?} First, we study the representative capability of the CLIP image feature space $\Omega_\text{I}$ by trying to generate 3D shapes from this space. Specifically, we replace the SVR image encoder $E_{\text{S}}$ %\xjqi{what is DVR for,a bit consufing, SVR and DVR is rather confusing. You should make this clear} 
with the CLIP image encoder $E_\text{I}$, and optimize implicit decoder $D$ using multi-view losses like DVR~\citep{niemeyer2020differentiable} with $E_{\text{I}}$ frozen. 
%In this way, we can replace $E_{\text{C\_I}}$ with CLIP text encoder $E_{\text{C\_T}}$ in inference for text-guided shape generation. 
This approach can be %easily \phil{NOTE: remove easily?} 
extended to text-to-shape generation by replacing $E_\text{I}$ with CLIP text encoder $E_\text{T}$ during the inference.
To compare the performance of $E_\text{S}$ and $E_\text{I}$, we evaluate 3D mIoU %mean Intersection over Union (mIoU)
between their generated shapes and GTs. The results are as follows: the standard SVR pipeline $E_\text{S}$+$D$ achieves 47.86\% mIoU while replacing the SVR encoder $E_\text{S}$ with CLIP encoder $E_\text{I}$ ($E_\text{I}$+$D$) degrades the performance to 40.09\%. %(\xjqi{it's difficult to understand what is ``$E_\text{S}$+$D$'' and ``xxxx'', can we have a small table. }).
From the results and qualitative comparison shown in Figures~\ref{fig:motivation} (a, b), we can see that the CLIP image space $\Omega_\text{I}$ has \textit{inferior representative capability to capture details of the input image} for 3D shape generation. This is not surprising, since the pre-trained $E_\text{I}$ from CLIP is targeted to extract semantic-aligned features from
%with \phil{with or from?}\lzz{with. from image, consistent with text} 
texts rather than extracting details from images.
%This is not surprising since $E_{\text{C\_I}}$ can represent a wide variety of categories with finite feature space dimensions; hence, it is challenging for CLIP feature to cover some details. On the contrary, $E_{\text{D}}$ only addresses on the $13$ categories and therefore can well capture details \xjqi{This explanation is not valid to me. A better explanation to me is that the pre-trained clip feature space is targeted to extract semantic features aligned to the text (some details relevant to 3D shape estimation such as xxxx might be lost). In contrast, the xxx space is optimized to xxxx therefore, necessary features xxxx??}. \xjqi{Highlight your conclusion, the clip feature space loses necessary details which motivate our first design xxxx.} 
Hence, image details relevant to 3D reconstruction are lost,~\eg, textures. On the contrary, $E_\text{S}$ from the SVR model is optimized for 3D generation from images, % \phil{from images,}
so it maintains more necessary details. The above result motivates us to design a mapper $M$ from $\Omega_\text{I}$ to $\Omega_\text{S}$ and then %\phil{then}
generate shapes 
%$S$ 
%\phil{shapes}
from $\Omega_\text{S}$ instead of $\Omega_\text{I}$ for better generative fidelity.

\vspace*{-3pt}
\subsubsection{How the CLIP image and text feature gap influences 3D shape generation?} 
%\subsubsection{How the unavoidable gap between CLIP image and text features influence 3D shape generation?} 
Second, we %study whether it is reasonable to directly replace $E_{\text{C\_I}}$ with $E_{\text{C\_T}}$ to enable the text input. To this end, we 
investigate the gap between the normalized CLIP image feature $f_\text{I} \in \Omega_\text{I}$ %$=E_\text{I}(I)/||E_\text{I}(I)||$ %(\xjqi{no need to have this equation, make it simpler, looks ugly})
and normalized CLIP text feature $f_\text{T}\in \Omega_\text{T}$;
%$=E_\text{T}(T)/||E_\text{T}(T)||$
%
(see also the CLIP image and text feature spaces
%\textcolor{purple}{as shown in 
in Figure~\ref{fig:figure1} (a))
%\lzz{change like this or remove the reference to Fig 1? } \phil{??? Fig.1a has been updated, so we need to update all related main texts in paper} 
and how such gap influences text-guided 3D shape generation. Specifically, %\xjqi{can we have some graphic illustration of this ??}
we randomly sample $300$ text-shape pairs from the text-shape dataset~\citep{chen2018text2shape}, then %compare the cosine distance
%\phil{between $f_\text{I}$ and $f_\text{T}$: when you say compare, there must be two parties},
evaluate the cosine distance between $f_\text{I}$ and $f_\text{T}$,~\ie, $d=1-\text{cosine\_similarity}(f_\text{I},f_\text{T})$, where $f_\text{I}$ is the CLIP feature of the rendered images from the corresponding shape. 
We repeat the experiment and obtain $d(f_\text{T},f_\text{I})=0.783\pm0.004$. %=0.2168\pm0.0041$. %and $\text{Sim}(f_T,f_I^u)=0.2022\pm0.0043$. 
The result reveals \emph{a certain gap between the CLIP text and image features in this dataset, even though they are paired.}
%with each other}. In addition,
Also, the angle in the feature space between the two features is around $\arccos(1-0.783)=1.35$ rad in this dataset~\citep{chen2018text2shape}. %\xjqi{why we need the similarity for the unparied data}. In addition, the similarity between the paired and randomly sampled $f_T,f_I$ are very close to each other (0.2022 $\vs$ 0.2168). 
Having said that, directly replacing $f_\text{I}$ with $f_\text{T}$ like~\citet{sanghi2021clip,zhou2021lafite} in inference may harm the consistency between the output shape and the input text.
 As demonstrated in Figure~\ref{fig:motivation} (c),  directly replacing $f_\text{I}$ with $f_\text{T}$ causes a cosine distance of 0.45 to $f_\text{S}\in \Omega_\text{S}$ (see Figure~\ref{fig:motivation} (c)), which is significantly larger than the distance between $M(f_\text{I})$ and $f_\text{S}$ (0.21). Our finding {is consistent with the findings in~\citet{liang2022mind}. It} motivates us to further fine-tune $M$ into $M'$ at test time, such that we can produce feature $M'(f_\text{T})$, which is closer to $f_\text{S}$ than $M(f_\text{T})$.
%\xjqi{For this part, how many samples you utilize to do this experiment, and how the different between text feature and image feature influence the final results. Please rethink about this}
%\phil{why test time? please add some more words to explain why here}\lzz{since "directly replacing $f_\text{I}$ with $f_\text{T}$ causes xxx cosine distance", then the only choice is test-time optimization. may remove "at test time" here?}
%\phil{Note: I tried to expand the last sentence}
%\phil{Note: better mention $M'$ in main text at least, as we have $M'$ in Figures 2 and 3}

%\xjqi{please summarize the conclusion and motivate the design of test-time optimization. The text feature is not aligned with the image feature even they are paired. explain why test-time optimization can work if text-image features can not be the same??} \lzz{the clip feature for a particular image has gap to text feature, but test-time optimization is not based on a particular image, it push text feature to go to the right direction. May not need to explain it? }

\vspace*{-5pt}

\subsection{Two-Stage Feature-Space Alignment}\label{sec:two-sgage}

Following the above findings, we propose a two-stage feature-space-alignment approach to 
first connect image space $\Omega_\text{I}$ and shape space $\Omega_{\text{S}}$ and further connect text space $\Omega_\text{T}$ to shape space $\Omega_{\text{S}}$ with the image space $\Omega_\text{I}$ as the stepping stone. 
%\phil{In paper writing, if we still have space, it is better to mention the full name before the math symbol. This will help readers to read the paper more smoothly, especially the symbol hasn't appeared for a while in the text.}
%\phil{approach to first connect $\Omega_\text{I}$ and $\Omega_{\text{S}}$ and further connect $\Omega_\text{T}$ to $\Omega_{\text{S}}$ with $\Omega_\text{I}$ as the stepping stone.}
%
%step-by-step connect $\Omega_\text{S},\Omega_{\text{I}}$, and $\Omega_{\text{T}}$.  %\xjqi{echo the previous two problems, and summarize what you will do with the two observations. In the following, we elaborate xxx which is xxxx. Then, we detail xxx to xxx.} 

\vspace*{-3pt}
\paragraph{Stage-1 alignment: CLIP image-to-shape mapping.}
%\xjqi{Given xxxx and xxx data, }
%\xjqi{could we find a better name for this? instead of two-stage alignment}
Given multi-view RGB or RGBD images for training, the stage-1 alignment is illustrated in Figure~\ref{fig:overview} (a).
%On the one hand, motivated by the finding of our first empirical study, it is desirable to generate shapes from $S_{SVR}$ instead of $S_{CLIP\_I}$ to ensure the high-fidelity generative results with fine details. On the other hand, CLIP offers joint text and image feature spaces that potentially enables text-free training.
Considering that shape space $\Omega_{\text{S}}$ contains richer object details than the image space $\Omega_{\text{I}}$, while $\Omega_{\text{I}}$ provides a joint text-image embedding with the input text space $\Omega_{\text{T}}$,
we introduce a fully-connected CLIP2Shape mapper $M$ to map image feature $f_{\text{I}}$ to shape space $\Omega_{\text{S}}$. Taking a rendered image $I$ as input, $M$ is optimized with an $L_2$ regression between $M(f_{\text{I}})$ and standard SVR feature $f_\text{{S}}=E_{\text{S}}(I)$ according to Equation~(\ref{equ:stage-1}) below:
\begin{equation}
\begin{aligned}
L_{M}=\sum^N_{i=1} ||M(f_{\text{I,i}})-E_{\text{S}}(I_i)||^2_2 
\label{equ:stage-1}
\end{aligned}
\end{equation}
where $N$ is the number of images in the training set %, \phil{no comma, when ``and'' is connecting only two items} 
and $f_{\text{I},i}$ is the normalized CLIP feature of $I_i$.

\begin{figure*}
\centering
\includegraphics[width=0.99\textwidth]{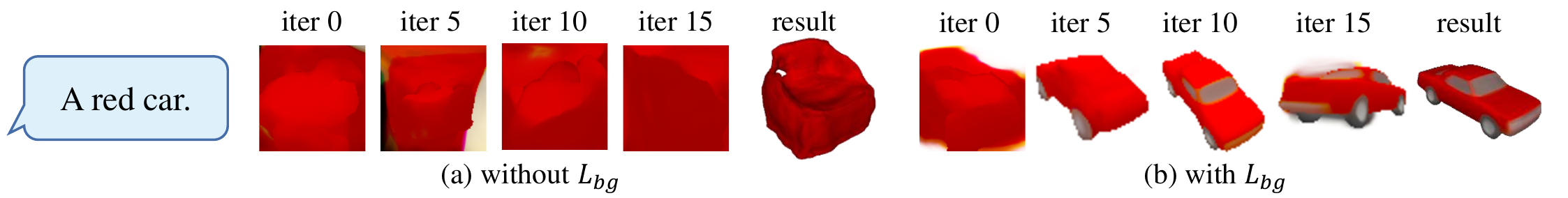}
\vspace*{-1.5mm}
\caption{Effect of generating shapes from the same text with/without background loss $L_\text{bg}$.
%\phil{where is this figure referenced? please move it to later pages? closer to the location that it is first referenced.}
%\caption{Effect of \phil{generating shapes from the same text with/without} background loss $L_\text{bg}$.
%\phil{where is this figure referenced? please move it to later pages? closer to the location that it is first referenced.}
}
% \xjqi{seems too small figures??} }
\label{fig:bg}
\vspace*{-2.5mm}
\end{figure*}

Also, we fine-tune decoder $D$ to encourage it to predict a white background, which helps the model to ignore the background and extract object-centric feature (see Figure~\ref{fig:bg}), while maintaining its 3D shape generation capability. %\xjqi{you fine-tuned D then its not forzen?? a bit weirded here}
To this end, we propose a new background loss $L_\text{{bg}}$ in Equation~(\ref{equ:bg}) below to enhance the model's foreground object awareness to prepare for the second-stage alignment.  %We will detail it later. 

%\phil{perhaps add one or two more sentences here to more explicitly explain why it is good to help the model to be more aware of the foreground object?}
\begin{equation}
\begin{aligned}
L_\text{bg}=\sum_p ||D_c(p)-1||_2^2 \mathbbm{1}(\text{ray}(o,p)\cap F =\emptyset)
\label{equ:bg}
\end{aligned}
\end{equation}
where $\mathbbm{1}$ is the indicator function;
$F=\{p: D_o(p)>t\}$ indicates the foreground region, in which the occupancy prediction $D_o(p)$ is larger than threshold $t$; $p$ is a query point;
$\text{ray}(o,p)\cap F =\emptyset$ means the ray from camera center $o$ through $p$ does not intersect the foreground object marked by $F$; and
$D_c(p)$ is the color prediction at query point $p$.
In a word, $L_{\text{bg}}$ encourages %\phil{who?} 
$D$ to predict the background region as white color (value 1), such that $E_\text{I}$ can focus on and better capture the foreground object.
% thanks
%} \phil{why this is useful... explain a bit more}.
In addition, to preserve the 3D shape generation capability of $D$, we follow the loss %es \phil{no es?} 
$L_{\text{D}}$, with which $D$ has been optimized in the SVR training.
In this work, we adopt~\citet{niemeyer2020differentiable}.

Hence, the overall loss in stage-1 training is $\lambda_{M}L_{M}$ for mapper $M$ and $\lambda_{\text{bg}}L_{\text{bg}}+L_{D}$ for decoder $D$, where $\lambda_{M}$ and $\lambda_{\text{bg}}$ are weights. 
The stage-1 alignment provides a good initialization for the test-time optimization of stage 2. 
%\phil{no hyphen for ''stage 2'' when it is used as a noun; add hyphen when it is used as an adj., e.g., ``stage-2'' alignment}
%\phil{so, please modify Fig.2 as well, Stage 1 rather than Stage-1 and Stage 2 rather than Stage-2 on left side of the figure}

%With stage-1 alignment, we are ready to replace $f_{CLIP\_I}$ with $f_{CLIP\_T}$ to enable text input, and 

\vspace*{-3pt}
\paragraph{Stage-2 alignment: text-to-shape optimization.}
%\xjqi{could we find a better name for this??}

%\begin{wrapfigure}{r}{0.5\textwidth}
%  \begin{center}
%    \includegraphics[width=0.48\textwidth]{figures/stage2.PNG}
%  \end{center}
%  \caption{Illustration of the stage-2 alignment. (a)-(d): rendered images in %training. (e): the generated shape. ``-'' minus ``without'' and ``+'' means ``with''. The input text is ``A long luxury black car''.}
%  \label{fig:stage-2}
%\end{wrapfigure}

Given a piece of text, the stage-2 alignment aims to further connect the text and shape modalities. Specifically, it searches for shape $S$ that best % \phil{best?}
matches the input text $T$. % by connecting the text and shape modalities \xjqi{connect ``text and shape'' is a bit weired and you use it many times}.
To this end, we formulate a fast test-time optimization to reduce the gap between the text and image CLIP features $f_{\text{T}}$ and $f_{\text{I}}$, as discussed earlier in the second empirical study. %, we suggest a fast test-time optimization as the stage-2 alignment. 

As shown in Figure~\ref{fig:overview} (b), given input text $T$, we replace image encoder $E_{\text{I}}$ with text encoder $E_{\text{T}}$ to extract CLIP text feature $f_{\text{T}}$, then fine-tune $M$ with CLIP consistency loss between input text $T$ and $m$ images $\{R_i\}_{i=1}^m$ rendered with random camera poses from output shape $S$; see Equation~\ref{equ:stage-2}:
%\phil{I removed $R$ as I guess you only need $\{R_i\}$; the rule is to define as few symbols as possible}
\begin{equation}
\begin{aligned}
L_{\text{C}}=\sum_{i=1}^m{\langle{f_\text{T}} \cdot \frac{E_{\text{I}}(R_i)}{||E_{\text{I}}(R_i)||}\rangle} %\cdot f_{\text{I},i}\rangle}
\label{equ:stage-2}
\end{aligned}
\end{equation}
%L_{M\_stage-2}= \sum_{i=1}^r \frac{\langle E_{CLIP\_I}(R_i),E_{CLIP\_{T}}(T)\rangle}{||E_{CLIP\_I}(R_i)||_2||E_{CLIP\_{T}}(T)||_2}
where $\langle\cdot\rangle$ indicates the inner-product. % and 
%$f_{\text{I},i}$ = $E_{\text{I}}(R_i)$. %, and $||\cdot||_2$ indicates the two-norm. 
%\phil{replace $f_{\text{I},i}$ with $E_{\text{T}}(R_i)$ in the equation?}

In stage-2 alignment, we still adopt $L_{\text{bg}}$ to enhance the model's foreground awareness.
Comparing Figures~\ref{fig:bg} (a) and (b), we can see that the stage-2 alignment is able to find a rough
%desired \phil{a rough?} 
shape with $L_\text{{bg}}$ in around %\phil{around} 
%less than 
five iterations, yet failing to produce a reasonable output without $L_\text{{bg}}$,
since having %\phil{having?}
the same color prediction on both foreground and background
%\lzz{not enforce. the background color tend to be same as foreground since there is no supervision on bg color. since the model tends to predict background to be the same color as foreground without $L_\text{bg}$}
%\lzz{due to the same color prediction of foreground and background 
hinders the object awareness of the model.
%\xjqi{please explain why your method can use much less time.} 
%\phil{need to explain more on how to interpret the figure before coming up with this conclusion}

Thanks to the joint text-image embedding of CLIP, the gap between text feature $f_\text{T}$ and shape feature $f_\text{S}$ has already been largely narrowed by $M$.
Therefore, 
%we 
%\phil{the stage-2 alignment}
the stage-2 alignment 
only needs to fine-tune $M$ with %only \phil{remove this only? too many only in this sentence} 
20 iterations using the input text, taking only around 85 seconds on a single GeForce RTX 3090 Ti, compared with 72 minutes taken by 
%which is much faster than the existing work 
Dream Fields~\citep{jain2021zero} at test time.
%which takes 72 minutes. 
After this fine-tuning, 
%Then, we can 
we can readily obtain a plausible result; see,~\eg, the ``result'' shown in Figure~\ref{fig:bg} (b). Our ISS is a novel and efficient approach for 3D shape generation from text.
%\phil{need to elaborate the quality of the result... as shown in Figure 4?}

%we only n fine-tune $M$ for 20 iterations for each input text, which takes around 85 seconds on a single GeForce RTX 3090 Ti, which is much faster than the existing work Dream Fields~\citep{jain2021zero} which takes 72 minutes. 

\vspace*{-3pt}
\paragraph{Diversified generation.}
In general, shape generation from text is one-to-many.
Hence, we further extend our approach
%Our approach can be readily extend for 
with diversified 3D shape generation from the same piece of input text.
Unlike the existing works, which require additional and complex modules,~\eg, GANs~\citep{chen2018text2shape}, IMLE~\citep{liu2022towards}, and normalizing flow network~\citep{sanghi2021clip}, we can simply perturb the image and text features for diversified generation. %\phil{can simply perturb the image and text features for diversified generation}.
%a few words to summarize how we did so here}.
%
Specifically, after stage-1 alignment, we randomly permute $f_{\text{I}}$ as an initialization and $f_{\text{T}}$ as the ground truth by adding normalized Gaussian noises $z_1=h_1/||h_1||, z_2=h_2/||h_2||$, where $h_1,h_2 \sim N(0,1)$ to derive diversified features
%$\hat{f}_{\text{I}}$ and $\hat{f}_{\text{T}}$:
%, respectively, as Equation~\ref{equ:random}. 
\begin{equation}
%\begin{aligned}
\hat{f}_{\text{I}}=\tau_1 f_{\text{I}}+(1-\tau_1)z_1 \ \ \text{and} \ \
\hat{f}_{\text{T}}=\tau_2 f_{\text{T}}+(1-\tau_2)z_2,
\label{equ:random}
%\end{aligned}
\end{equation}
where $\tau_1,\tau_2$ are hyperparameters %\phil{hyper-parameters or hyperparameters? Which one is more common?}
to control the degrees of permutation. With permuted $\hat{f}_{\text{I}}$ and $\hat{f}_{\text{T}}$ in stage-2 alignment, our model can converge to different 3D shapes for different noise. 
%each time. 

\vspace*{-3pt}
\subsection{Text-Guided Stylization}\label{sec:stylization}

%\xjqi{How this is related to other modules}
The two-stage feature-space alignment is already able to generate plausible 3D shapes; see,~\eg,
%those shown in Figure~\ref{fig:figure1} (c) and 
Figures~\ref{fig:overview} (b) and~\ref{fig:bg} (b).
However, the generative space is limited by the representation capability of the employed
%pre-trained 
SVR model,~\eg,~DVR~\citep{niemeyer2020differentiable} can only generate shapes with limited synthetic %\textcolor{blue}{structure and texture} 
patterns as those in ShapeNet. % \phil{shapes with limited synthetic texture patterns as those in ShapeNet}.
However, 
a richer and wider range of 
%more fascinating and diversified \phil{these words have been used when you talk about the diversified shape generation, so it may confuse people with ``diversified shape generation'' and this ``text-guided stylization'' here}
structures and textures are highly desired. To this end, we equip our model with a text-guided stylization module %\pdai{, without manually designing the surface material,}
to enhance the generated shapes with novel structure and texture appearances, as shown in Figures~\ref{fig:overview} (c) and~\ref{fig:figure1}.

Specifically, for texture stylization, we first duplicate $D$ (except for the output layer) to be $D_o$ and $D_c$, then put the output occupancy prediction layer and color prediction layer on top of $D_o$ and $D_c$, respectively.
%. $D_o$, with a single output channel, inherits the occupancy prediction branch from $D$, and $D_c$ with three output channels inherits the color prediction. 
Further, we fine-tune $D_c$ with the same CLIP consistency loss as in Equation~(\ref{equ:stage-2}), encouraging the consistency between input text $T$ and the $m$ rendered images $\{R_i\}_{i=1}^m$.
%$R=\{R_1, \dots, R_m\}$ 

%\textcolor{purple}{
Besides textures, novel structures are also desirable
%significant 
for shape stylization. Hence, we further incorporate a shape-and-texture stylization strategy to create novel structures.
To enable 
%\phil{encourage?} \lzz{I think enable is better, since the previous mentioned "texture stylization" cannot change shape. }
shape sculpting, we fine-tune $D$ with the same CLIP consistency loss in Equation~\ref{equ:stage-2}. At the same time, to maintain the overall structure of the initial shape $S$, we propose a 3D prior loss $L_P$ shown in Equation~(\ref{equ:prior}), aiming at preserving the 3D shape prior learned by the two-stage feature-space alignment. 
\begin{equation}
%\begin{aligned}
L_\text{P}=\sum_p |D_o(p)-D_o^\prime(p)|
%\end{aligned}
\label{equ:prior}
\end{equation}
where $p$ is the query point, and $D_o$, $D_o^\prime$ are the occupancy predictions of the initial $D$ and the fine-tuned $D$ in the stylization process, respectively.
%} 
%
To improve the consistency between the generated texture and the generated shape, we augment the background color of $R_i$ with a random $\text{RGB}$ value in each iteration.
Please find more details in the supplementary material.

\vspace*{-6pt}
\subsection{Compatibility with Different SVR models}
%\subsection{Broaden Shape Categories with Different SVR models}
%\subsection{Extend to a Wide Range of Categories}
\label{sec:compatible} 

%\lzz{may change the name of this subsection to be "Compatibility with Different SVR models?". }
%\phil{good suggestion}

Besides DVR~\citep{niemeyer2020differentiable}, our ISS framework is compatible with different SVR models.
For example, we can adapt it with the most recent SVR approach SS3D~\citep{alwala2022pre} that leverages in-the-wild single images for 3D generation.
With this model, our framework can generate a wider %\phil{wider?} 
range of shape %\phil{shape}
categories 
%by replacing \phil{shape encoder} $E_{\text{S}}$ and \phil{decoder} $D$ with SS3D's encoder $E_{\text{SS3D}}$ and decoder 
by using SS3D's encoder and decoder as shape encoder $E_{\text{S}}$ and decoder $D$ in our framework, respectively. Here, we simply follow the same pipeline as in Figure~\ref{fig:overview} to derive a text-guided shape generation model for the in-the-wild categories; see our results in Section~\ref{sec:exp_ss3d}.
%\lzz{as shown in Figure~\ref{fig:ss3d}}.
Notably, we follow the losses in~\citet{alwala2022pre} in place of $L_\text{D}$ (see Section~\ref{sec:two-sgage}) in stage-1 training, requiring only single-view images without camera poses. 
More importantly, our approach's high compatibility suggests that it is orthogonal to SVR, so its performance can potentially be further upgraded with more advanced SVR approaches in the future.

%\phil{do you need to define $E_{\text{SS3D}}$ and $D_{\text{SS3D}}$?  If we do not need them later on, remove these two symbols} \lzz{Thanks. Have removed. }

%\phil{I assume that you will show some results produced with SS3D and some with DVR later, right?  Mention here?  This subsection only states that our framework is compatible with different SVR models but does not provide any visual evidence that it really can work with different SVR models}

%to illustrate how our approach works with general SVR models. 

%Specifically, we replace the 

\vspace*{-10pt}

\section{Experiments}
\label{sec:results}

%In this section, we first introduce the employed datasets, the implementation details, and the evaluation metrics in Sections~\ref{sec:implementation}.
%Then present ablation studies in Section~\ref{sec:ablation}.
%Further, we study the generative novelty and diversity, scalability, and generality of our ISS approach in Section~\ref{sec:show}.

\vspace*{-7pt}
\subsection{Datasets, Implementation Details, and Metrics}~\label{sec:implementation}
%\subsection{Datasets and Implementation Details}~\label{sec:implementation}

\vspace*{-12pt}
%\paragraph{Datasets}
With multi-view RGB or RGBD images and camera poses, we can train ISS on the synthetic dataset ShapeNet~\citep{shapenet2015} (13 categories) and the real-world dataset CO3D~\citep{reizenstein2021common} (50 categories). To evaluate our generative performance, we create a text description set with four texts per category on ShapeNet and two texts per category on CO3D. 
SS3D~\citep{alwala2022pre} takes single-view in-the-wild images in training; as their data has not been released, % \xjqi{this sentence is difficult to digest, what do you mean by ``with single view in-the-wild images''}; 
we only evaluate our method on some of their categories. %on a subset of them including PASCAL3D+~\citep{xiang2014beyond} containing 11 rigid-categories and CUB-200-2011~\citep{wah2011caltech}. 
%Implementation details are provided in the supplementary material. %
%Metrics are introduced in the supplementary material. 
To evaluate the performance, we employ Fréchet Inception Distance (FID)~\citep{heusel2017gans}, Fréchet Point Distance (FPD)~\citep{liu2022towards} to measure shape generation quality, and conduct a human perceptual evaluation to further assess text-shape consistency. Please refer to the supplementary material for more details on the metrics and implementation details.

\vspace*{-10pt}
\subsection{Comparisons with State-of-the-art Methods}\label{sec:existing}
\begin{figure*}[!t]
\centering
\includegraphics[width=0.99\textwidth]{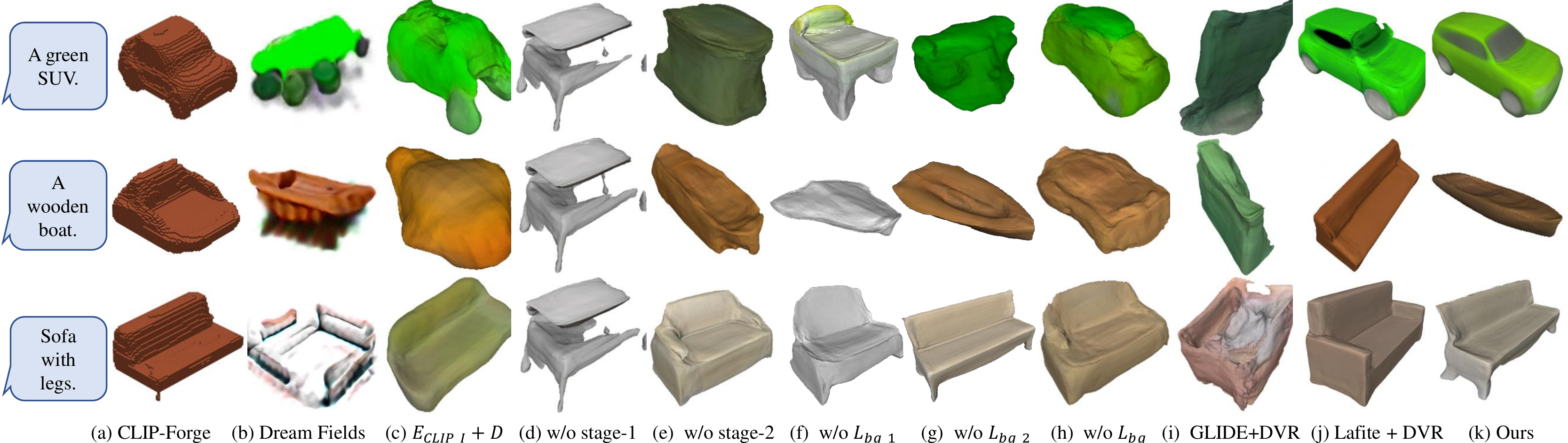}
\vspace*{-1.5mm}
\caption{Qualitative comparisons with existing works and baselines.
%\phil{try to better center align the elements in each column}
}
\label{fig:quality}
\vspace*{-2.5mm}
\end{figure*}

We compare our approach with existing works~\citep{sanghi2021clip,jain2021zero} both qualitatively and quantitatively. For a fair comparison, we use their official codes on GitHub to generate shapes on our text set.
Table~\ref{tab:all} shows quantitative comparisons, whereas Figure~\ref{fig:quality} shows the qualitative comparisons.
Comparing existing works and ours in Table~\ref{tab:all},
%results shown in Table~\ref{tab:all} (``Existing works'' and ``Ours'') and Figure~\ref{fig:quality} (a,b,i), 
we can see that our approach outperforms two state-of-the-art works by a large margin for both generative quality and text-shape consistency scores. %\phil{scores}.
On the other hand, the qualitative comparisons in Figure~\ref{fig:quality} (a,b) show that CLIP-Forge~\citep{sanghi2021clip} produces low-resolution shapes without texture, and some generated shapes are inconsistent with the input text,~\eg, ``a wooden boat.''
Dream Fields~\citep{jain2021zero} cannot 
%fails to 
generate reasonable shapes from the input text on the top row and its generated shape on the bottom row is also inconsistent with the associated input text.
On the contrary, our approach (Figure~\ref{fig:quality} (i)) can generate high-fidelity shapes that better match the input text. Note that we only utilize two-stage feature-space alignment without stylization in producing our results. Please refer to the supplementary file for more results and visual comparisons. %\phil{and visual comparisons?}.

\begin{comment}
\begin{table}
\centering
\caption{A/B test results and Consistent Score of LAFITE+DVR and ours}
\label{tab:ab}
\vspace*{-1mm}
\scalebox{1}{
  \begin{tabular}{ccc}
    \toprule
    Method & Number of choice ($\uparrow$)  & Consistency Score (\%) ($\uparrow$)   \\
    \midrule
     LAFITE~\citep{zhou2021lafite}+DVR~\citep{niemeyer2020differentiable} & 15.4$\pm$4.69 & 61.15$\pm$11.13  \\
    \midrule
    Ours & \textbf{20.90$\pm$5.91} & \textbf{66.73$\pm$9.51}   \\
    \bottomrule
  \end{tabular}
  }
\end{table}    
\end{comment}

\begin{comment}
\begin{figure}[htb] 
  \begin{minipage}[b]{0.7\textwidth} 
  \end{minipage} 
  \begin{minipage}[b]{0.25\textwidth} 
    \centering 
    \includegraphics[width=0.25\textwidth]{diversified.pdf}
    \caption{This is a Figure by a Table} 
    \label{fig:by:table} 
  \end{minipage}% 
\end{figure}
\end{comment}

\vspace*{-10pt}

\subsection{Ablation Studies}\label{sec:ablation}

To manifest the effectiveness of our approach, we conduct ablation studies on the following baselines (see Table~\ref{tab:all} and Figure~\ref{fig:quality}): generate shapes from $\Omega_I$ ($E_{\text{I}}+D$), optimize stage-2 alignment without stage-1 (w/o stage 1), conduct stage-1 alignment without stage-2 (w/o stage 2), disable the background loss in stage 1, stage 2 and both (w/o $L_{\text{bg\_1}}$, w/o  $L_{\text{bg\_2}}$, w/o  $L_{\text{bg}}$), and two additional baselines that first create images and then 3D shapes (GLIDE+DVR, LAFITE+DVR). More details on the setup and analysis are provided in the supplementary material Section 3. Our results outperform all the existing works and baselines in terms of fidelity and text-shape consistency by a large margin. %\xjqi{provide details about the figure??}.

\vspace*{-5pt}

\subsection{More Analysis on Generative Results of ISS} \label{sec:show}
\vspace*{-5pt}
% \xjqi{put this part in the main results session, move before ablation studies would be better}

%\xjqi{add a summary of what will be detailed in the following. What is the purpose of this part. summarize some key properties of ISS: Scalability, generality xxx}
Next, we present evaluations on the generative novelty and diversity, as well as the scalability of our two-stage feature-space alignment. Then, we show more text-guided stylization results and how our ISS approach generalizes to a wide range of categories and generates shapes with better fidelity.

%CLIP-Mesh: FID 188.09200326730937, consistency score, 32.789 +- 8.82
%CLIP-Mesh:val IS:  (2.5366502, 0.22608429)
%CLIP-Mesh:pred val fid:  (37.612865331328614-2.512834739152946e-07j)

%Div FID 108.73
%Div IS: 3.80
%Div fid: 31.92
\begin{table*}[!t]
\centering
\caption{Quantitative comparisons with existing works and baselines.}
%\vspace*{-1mm}
\scalebox{0.81}{
  \begin{tabular}{ccccccc}
    \toprule
   Method Type & Method & FID ($\downarrow$) & Consistency Score (\%) ($\uparrow$) & FPD ($\downarrow$) & A/B/C Test  \\
    \midrule
    \multirow{2}{*}{\rotatebox{0}{
Existing works
}}&  CLIP-Forge  & 162.87 %173.77
&  41.83 $\pm$ 17.62 & 37.43 & 8.90 $\pm$ 4.12  \\
    & Dream Fields & 181.25 &  25.38 $\pm$ 12.33  & N.A. & N.A. \\
    \midrule
    \multirow{6}{*}{\rotatebox{0}{
Ablation Studies
}}  & $E_{\text{I}}$+$D$ & 181.88 & 20.97 $\pm$ 13.59 & 38.61 & N.A. \\
    & w/o Stage 1 & 222.96 & 1.92 $\pm$ 2.22  & 79.41 &  N.A.  \\
    & w/o Stage 2 & 202.33 & 29.52 $\pm$ 14.86  & 41.71 &  N.A.\\
    & w/o $L_{\text{bg\_1}}$ & 149.45 & 29.45 $\pm$ 14.67 & 40.85 & N.A. \\
    & w/o $L_{\text{bg\_2}}$ & 156.52  & 31.55 $\pm$ 8.87 & 38.31 & N.A. \\
    & w/o $L_{\text{bg}}$ & 178.34 & 30.96 $\pm$ 15.49 & 40.98 & N.A. \\
    \midrule
    \multirow{2}{*}{\rotatebox{0}{
Text2Image+SVR
}}  & GLIDE+DVR & 212.41 & 8.85 $\pm$ 7.94 & 41.33 & N.A. \\
    & LAFITE+DVR & 135.01 & 52.12 $\pm$ 11.05  & 37.55 & 11.70 $\pm$ 4.11\\
    \midrule
    Ours & ISS & \textbf{124.42 $\pm$ 5.11} & \textbf{60.0 $\pm$ 10.94}  & \textbf{35.67 $\pm$ 1.09} & \textbf{21.70 $\pm$ 5.19}  \\
    \bottomrule
  \end{tabular}
  }
  \label{tab:all}
\end{table*}    

\begin{figure*}[!t]
\centering
\includegraphics[width=0.99\textwidth]{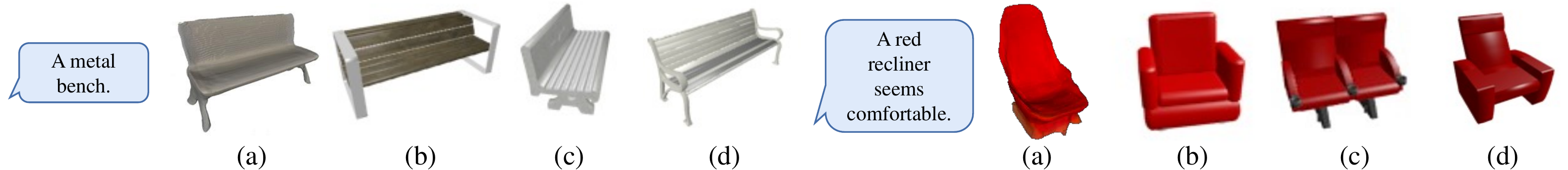}
\vspace*{-3.5mm}
\caption{Our approach is able to generate novel shapes, not in the training set.
(a) shows our results and (b,c,d) are the top-three shapes retrieved from the training set.
%\phil{TODO: center align the text labels with the image above each of them}
}
\label{fig:novelty}
\vspace*{-3.5mm}
\end{figure*}

\begin{figure*}[!t]
\centering
\includegraphics[width=0.9\textwidth]{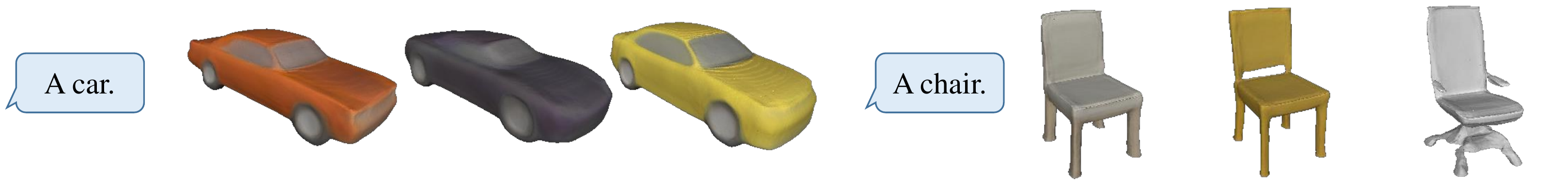}
\vspace*{-3.5mm}
\caption{Our approach can generate diversified results from the same input text.}
%given one text. }
\label{fig:diversified}
\vspace*{-3.5mm}
\end{figure*}

\begin{figure*}[!t]
\centering
\includegraphics[width=0.99\textwidth]{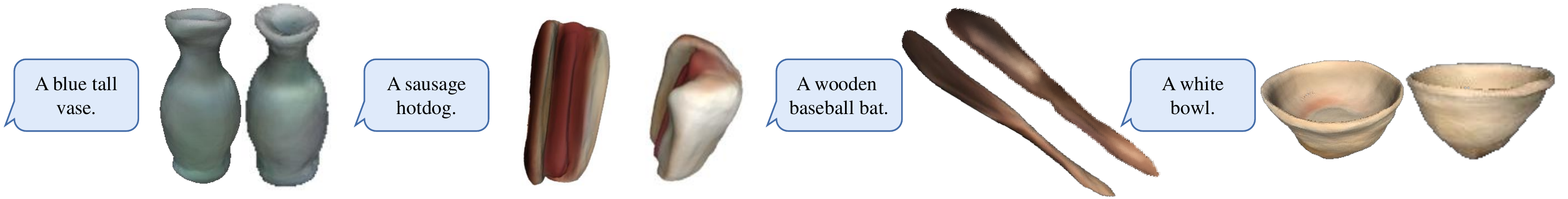}
\vspace*{-3.5mm}
\caption{Results on CO3D dataset. We show two different views of each result.}
%we show rendered images from two different views. %\phil{this figure is never referenced}
%}
\label{fig:co3d}
\vspace*{-3.5mm}
\end{figure*}

\vspace*{-2.5mm}
\begin{figure*}[!t]
\centering
\includegraphics[width=0.99\textwidth]{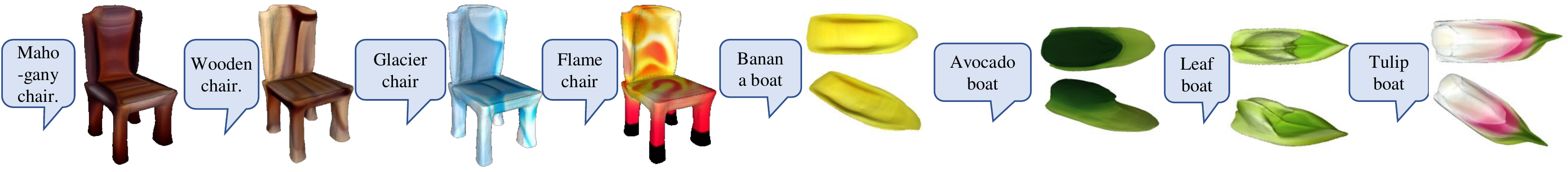}
\vspace*{-2.5mm}
\caption{Text-guided stylization. Left: texture stylization. Right: shape-and-texture stylization.}
\label{fig:style}
\vspace*{-2.5mm}
\end{figure*}

\begin{figure*}[!t]
\centering
\includegraphics[width=0.99\textwidth]{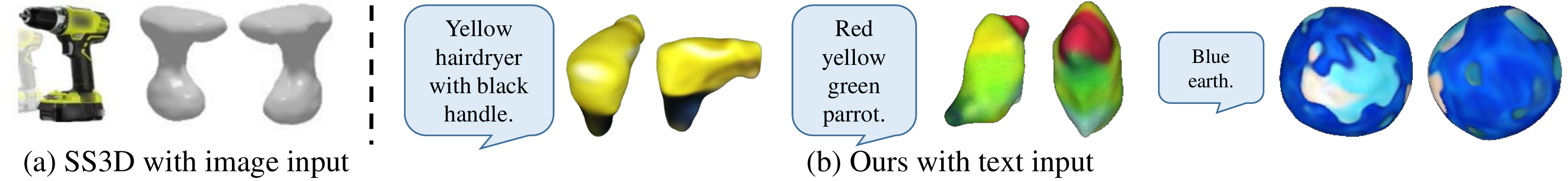}
\vspace*{-2.5mm}
\caption{With single images for training (without camera poses), our approach can produce results for a broad range of categories, by adopting~\citep{alwala2022pre}. Two different views are rendered. } % for each result.}
%shown We show rendered images from two different views. }
\label{fig:ss3d}
\vspace*{-3.5mm}
\end{figure*}

\vspace*{-3.5mm}

\begin{figure*}
\centering
\includegraphics[width=0.99\textwidth]{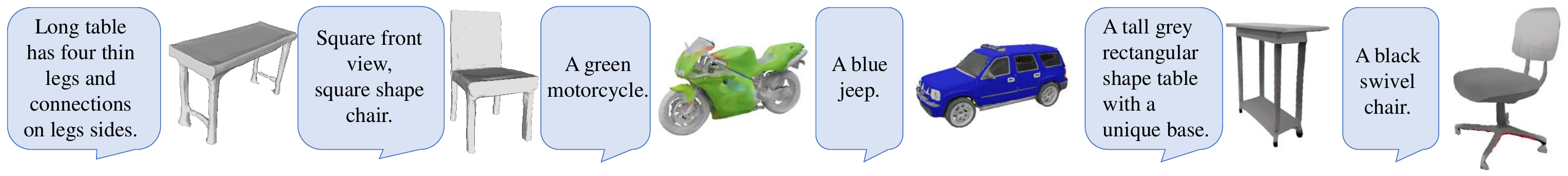}
\vspace*{-3.5mm}
\caption{Results of ISS built upon the SVR model of IM-Net (left two) and GET3D (right four).}
\label{fig:im-net}
\vspace*{-3.5mm}
\end{figure*}

\vspace*{0.2in}
\paragraph{Generation novelty.}
Our approach is able to generate novel shapes beyond simple retrieval from the training data. As shown in Figure~\ref{fig:novelty}, from the input text, we first generate our result in (a) and then take our generated shape to retrieve the top-three closest shapes in %\phil{from the input text, we first generate our result in (a) and then take our generated shape to retrieve the top-three closest shapes in} 
%we present top-three retrieved shapes in 
the %\phil{associated} 
associated training set based on the cosine similarity between the CLIP features of the rendered images $f_\text{I}$ and input text $f_\text{T}$ as the retrieval metric.
%
%cosine similarity between the rendered images and input text as the retrieval metric. 
%\phil{this clause is not clear: give a bit more details on how to compute the cosine similarity}
%
Our results with the two-stage feature-space alignment are not in the training sets,
%\phil{are not in the training sets},
%are different from the shapes in training sets, 
showing that our ISS approach can generate novel shapes beyond the training set, even without stylization.
It is not surprising, as our approach shares the generative space with the pre-trained SVR model and can potentially generate all shapes that the pre-trained SVR model can produce.

%\vspace*{-3pt}
\paragraph{Generation diversity.}
%{Equipped with feature perturbation module}\lzz{do you think this is a module?} 
By perturbing the features and injecting randomness on initialization, ISS is able to generate diversified results from the same text input. As shown in Figure~\ref{fig:diversified}, ISS produces various cars and chairs from the same piece of text. Quantitative results are presented in the supplementary material Section 3.4 {where our method achieves a high PS (Point Score).} 
%\xjqi{provide more details about which section in th}.

\vspace*{-6pt}
\paragraph{Generation fidelity.}

To assess the capability of ISS in generating realistic real-world 3D shapes, we train the SVR model on the CO3D dataset~\citep{reizenstein2021common} which is a real-world dataset, and ISS leverages the learned feature space for text to shape generation without paired data. As shown in Figure~\ref{fig:co3d}, our model is able to generate real-world shapes. As far as know, this is first work that investigates text-guided shape generation and on real-world datasets can generate realistic 3D shapes.
%in Figure~\ref{fig:co3d}.  he empolyed SVR model is trained on this dataset. ISS is then trained to leverage the trained SVR model for 3D text-shape generation.   As far as we know, this is the first work that investigates real-world text-guided 3D shape generation. \xjqi{Our method achieves 

%, so we just report that we achieve 124.31 FID on CO3D dataset, as a baseline for future works. 

\vspace*{-6pt}
\paragraph{Generation beyond the capability of SVR model.} 
Our text-guided stylization module equips our model with the capability to generate 3D shapes beyond the SVR model. As shown in Figure~\ref{fig:style} and Figure~\ref{fig:overview} (c), our model is able to create realistic and fantasy novel structures and textures that match text descriptions. Please refer to the supplementary material Section 4 for more details.

%More text-guided stylization results are shown in Figure~\ref{fig:style}, in addition to Figure~\ref{fig:figure1} (b) and Figure~\ref{fig:overview} (c). Our text-guided stylization approach is able to create both realistic and fantasy novel structures and textures \lzz{beyond the capability of SVR model. Please refer to the supplementary material Section 4 for more details.} % that do not exist in the training data. \xjqi{supplementary ??}   

\vspace*{-6pt}
\paragraph{Generality and Scalability of ISS on other SVR models.}\label{sec:exp_ss3d}
Our model is generic and can work together with other SVR models. To evaluate the generality and scalablity of our model, we employ SS3D~\citep{alwala2022pre}, IM-Net~\citep{chen2019learning}, and GET3D~\citep{gao2022get3d} as  SVR models to provide the feature space. It is worth noting that SS3D is capable of generating shapes of more categories and IM-Net, GET3D can produce high fidelity results. First, as shown in Figure~\ref{fig:ss3d}, built upon SS3D,  our approach can generate shapes of more real-world categories,~\eg, bird. 
%As shown in Figure~\ref{fig:ss3d} (a,b), given text descriptions, 
Note that our model %with  %\phil{only?}
% text as the input 
can generate shapes with comparable or even better qualities compared with initial SS3D model that takes an image as input.
%Besides, our text-guided texture stylization modules further allows the model to generate novel textures well-aligned with shapes. \xjqi{This seems do not have much relation with text-guided stylization. In the figure, you'd better add some explanation that (c) is obtained with text-guided stylization.} (see Figure~\ref{fig:ss3d} (c)). 
%it \phil{who? which figure? which result?} \lzz{Figure~\ref{fig:ss3d} (c)} 
%Figure~\ref{fig:ss3d} (c) further manifests that our stylization module can generate textures that are well-consistent with the shape.
%\xjqi{move the training details to supplementary}.
Second, when combined with IM-Net and GET3D, our model can fully exploit their generative capabilities and produces high-quality 3D shapes as shown in Figure~\ref{fig:im-net}.
The above manifests that ISS is generic and compatible to advanced models for generating shapes of more categories and higher qualities.

\vspace*{-3.5mm}

\section{Conclusion}

In this paper, we present a novel approach for text-guided 3D shape generation by leveraging the image modality as a stepping stone.  Leveraging the joint text-image embeddings from CLIP and 3D shape priors from a pre-trained SVR model, our approach eliminates the need for the paired text and shape data.
Technically, we have the following contributions.
First, we step-by-step reduce the semantic gap among the text, image and shape modalities through our two-stage feature-space alignment approach.
Second, our text-guided stylization technique effectively enriches our generated shapes with novel structures and textures in various styles.
Third, our approach is compatible with various single-view reconstruction approaches and can be further extended to generate a wide range of categories with only single images without camera poses in training.
Experiments on ShapeNet, CO3D, and multiple single-image categories manifest the superiority of our framework over the two state-of-the-art methods and various baselines.
Limitations are discussed in the supplementary files.

\section*{ACKNOWLEDGEMENTS}
The work has been supported in part by  the Research Grants Council of the Hong Kong Special Administrative Region (Project no. CUHK 14206320), General Research Fund of Hong Kong (No. 17202422), Hong Kong Research Grant Council - Early
Career Scheme (Grant No. 27209621), and National Natural Science Foundation of China (No. 62202151).

%\bibliography{iclr2023_conference}
%\bibliographystyle{iclr2023_conference}

%\appendix
%\section{Appendix}
%You may include other additional sections here.

%\clearpage
%\appendix
\vspace{+20mm}
\centerline{\Large{\textbf{Supplementary Material}}}
In this supplementary material, we first introduce the background augmentation in text-guided shape stylization (Section~\ref{sec:back_aug_supp}). Next, we present the implementation details and evaluation metrics, and provide the details of human perceptual evaluation results
(Section~\ref{sec:implement}). Then, we introduce the setup and analysis of our ablation studies (Section~\ref{sec:ablation_supp}). %Also, we provide our generative results with a stronger SVR model of IM-Net~\citep{chen2019learning} (Section \label{im-net}).
After that, we discuss text-guided stylization and provide more results  (Section~\ref{sec:stylization}). Then we show additional generative results of our approach (Section~\ref{sec:results}). Further, we discuss two alternative training strategies (Section~\ref{sec:alternative}). Then, we show how feature is mapped in the latent space  (Section~\ref{sec:mapping}) and also review some related literatures about SVR and differentiable rendering  (Section~\ref{sec:review}). Afterwards we compare the number of parameters in our model and existing works  (Section~\ref{sec:parameter}). We also discuss failure cases (Section~\ref{sec:failure}) and limitations of this work (Section~\ref{sec:limitation}). Finally, we provide our text  sets (Section~\ref{sec:textset}) and summarize the notations in the paper (Section~\ref{sec:notations}).

\section{Background Augmentation in Text-Guided Shape Stylization}\label{sec:back_aug_supp}

% the existing works that focus on text-guided image generation~\citep{ramesh2021zero,zhou2021lafite,wang2022clip,ramesh2022hierarchical}, our text-guided shape stylization has an additional requirement that the generated textures should be consistent with the given shape. 

\begin{figure*}
\centering
\includegraphics[width=0.99\textwidth]{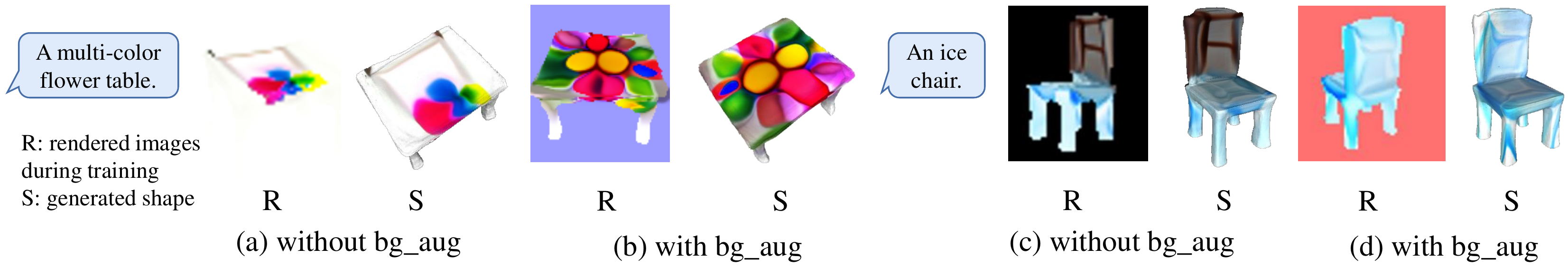}
\vspace*{-1.5mm}
\caption{Effective of text-guided shape stylization with/without background augmentation.
}
\label{fig:bg_aug_supp}
\vspace*{-2.5mm}
\end{figure*}

One important thing in text-guided shape stylization is that the generated texture should align with the given shape. However, it cannot be ensured with a simple
white or black background during training since the generated textures can be affected by the background color. As shown in Figure~\ref{fig:bg_aug_supp} (a), the shape in white color may confuse with the white background; thus, the model would struggle to capture the boundary of objects, hence cannot generate textures that well-align with the table. In Figure~\ref{fig:bg_aug_supp} (c), the generated texture is severely affected by the black background color, causing low-quality stylization results. 

To address the above issue, we propose a background augmentation strategy to improve the alignment between texture and shape. Specifically, the background color is replaced with random RGB values in each training iteration. By this means, the foreground shape may be more easily captured in training as shown in Figure~\ref{fig:bg_aug_supp} (b,d), improving the texture-shape consistency and the stylization quality.

\paragraph{Discussion on $L_\text{bg}$ and background augmentation.}

\begin{figure*}
\centering
\includegraphics[width=0.99\textwidth]{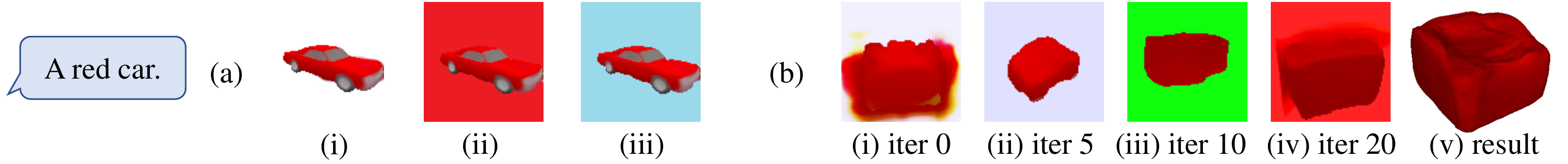}
\vspace*{-1.5mm}
\caption{An investigation on background loss and background augmentation. (a) Background color affects the cosine similarity of the CLIP features between the image and the text ``a red car'', \ie, (i) 0.292 (ii) 0.303 and (iii) 0.285. (b) Effect of generating shapes with background augmentation, but without background loss $L_{\text{bg}}$. Comparing with Figure 4 (b) in the main paper, the two-stage feature space alignment works well with $L_{\text{bg}}$, but fails with the background augmentation. 
}
\label{fig:bg_aug_supp2}
\vspace*{-2.5mm}
\end{figure*}

In two-stage feature space alignment (Section 3.3 in the main paper), we introduce a background loss $L_\text{bg}$ to encourage the color prediction on the background region to be white. A natural question is whether we can use background augmentation as a replacement, and our answer is no. As shown in Figure~\ref{fig:bg_aug_supp2} (a), the background color can affect the cosine similarity of CLIP features between the image and input text; %\xjqi{what if you want to generate a shape with white color?? Say sth about the white color is xxx we find that has the least chance to be confused with xxx and at the same time facilitate stability??} 
thus, using different background color in each iteration makes stage-2 alignment unstable and affects shape generation. As a result, the two-stage alignment can only benefit from $L_{bg}$, but not from the background augmentation, for producing a plausible shape. 
Besides, we empirically found that the two-stage feature space alignment with $L_\text{bg}$ performs well, even if a white shape is being considered, see the bottom row in Figure~\ref{fig:quality_supp} (i).

%On the other hand, shape stylization benefits from the background augmentation, since only texture is updated with the shape fixed in the stylization process, which makes stylization an easier task compared with the two-stage alignment, and can tolerate feature variations with different background colors. 

\section{Implementation Details, Metrics, and Human Perceptual Evaluation Details}~\label{sec:implement}
\paragraph{Implementation details.}
Our framework is implemented using PyTorch~\citep{paszke2019pytorch}. We first train the stage-1 CLIP-image-to-shape mapping %\xjqi{modify stage-1} 
for 400 epochs with learning rate $1e^{-4}$, and then train 
%\xjqi{modify stage-2} 
the stage-2 text-to-shape module at test time for 20 iterations which takes only $85$ seconds on average on a single GeForce RTX 3090 Ti.
Optionally, we can further train text-guided stylization with the same learning rate.
We %\phil{empirically}
empirically set hyperparameters $\lambda_{M}$, $\lambda_{bg}$, $t$, $m$, $\tau_1$, $\tau_2$ to be 0.5, 10, 0.5, 10, 0.2, 0.95, respectively, according to a small validation set.

\paragraph{Details on camera poses.}

In Stage 1, we follow~\citet{niemeyer2020differentiable} to set the camera poses to encourage the background to be white. Specifically, we randomly sample the distance of the camera and the viewpoint on the northern hemisphere. 

In Stage 2, compared with~\citet{niemeyer2020differentiable}, we sample the camera distance to be 1.5 times further compared with~\citet{niemeyer2020differentiable}. It helps to encourage sampling more global views instead of only local ones, so that the CLIP image encoder can capture the whole shape and yield a better CLIP feature.

In Stage 3, we also sample the camera distance to be 1.5 times. Since this stage aims to generate textures instead of searching for a target shape like Stages 1 and 2, only sampling view points on the northern hemisphere of the view space cannot ensure good generation quality in the bottom regions. Thus, we further randomly sample viewpoints on the southern hemisphere for random 10\% training iterations to encourage the stylized results to be consistent with the text in various viewpoints.

\paragraph{Decoder duplication}
Except for the output layer, we simply duplicate them to be $D_o$ and $D_c$. For the output layer, it takes $d$-dimensional features as input and outputs one value for occupancy and three values for RGB. We then copy $d\times1$ weights to $D_o$ and copy $d\times3$ weights to $D_c$. See Figure~\ref{fig:dup} for more details.

\vspace*{-3pt}
\paragraph{Metric: Shape generation quality.}
To measure the shape generation quality, we employ Fréchet Inception Distance (FID)~\citep{heusel2017gans} between five rendered images of the generated shape with different camera poses and a set of ground-truth ShapeNet or CO3D images. We adopt the official model with Inception Net trained on ImageNet, which is widely used to evaluate generative quality and realism. We do not train a model on ShapeNet, since it is too small to train a better network for evaluating the FID than models trained on ImageNet.
In addition, we randomly sample 2600 images in the ShapeNet dataset as ground truths for FID evaluation, instead of using images from ImageNet. It helps to measure the similarity between the generated shapes and ground truths of ShapeNet.

\begin{figure*}
\centering
\includegraphics[width=0.7\textwidth]{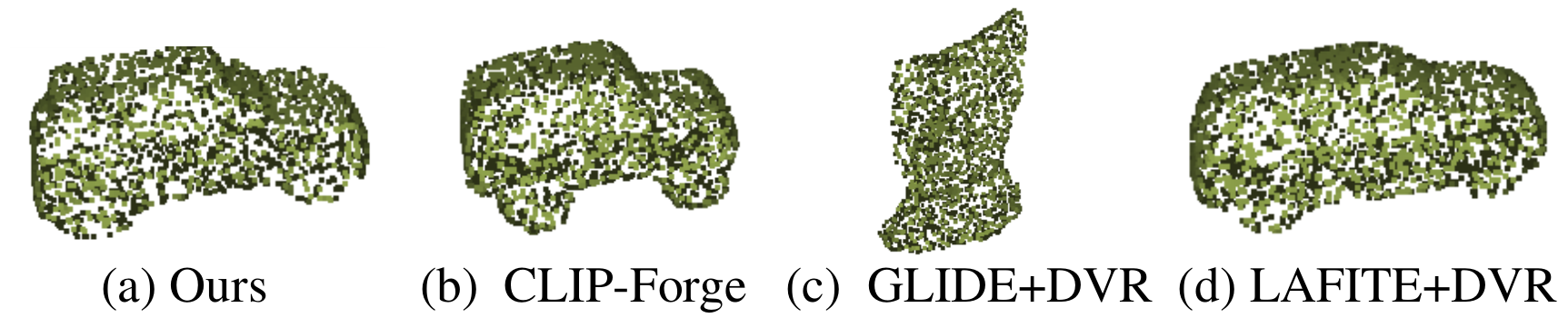}
\vspace*{-1.5mm}
\caption{Visualization of point clouds of different methods for FPD evaluation.
}
\label{fig:point_cloud}
\vspace*{-2.5mm}
\end{figure*}

\begin{figure*}
\centering
\includegraphics[width=0.99\textwidth]{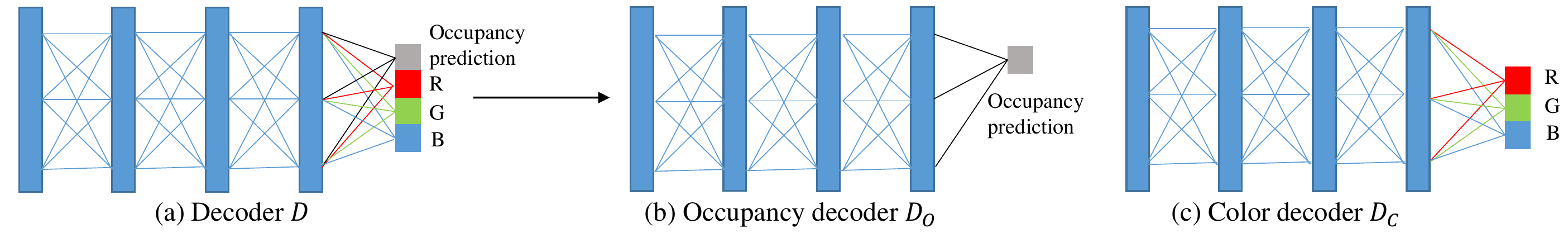}
\vspace*{-3.5mm}
\caption{Visualization to duplicate the decoder $D$ to $D_o$, $D_c$.}
\label{fig:dup}
\vspace*{-3.5mm}
\end{figure*}

Besides adopting FID, we further utilize the metric Fréchet Point Distance (FPD) proposed in~\citep{liu2022towards} to measure the shape generation quality without texture. We first convert the generated shapes to 3D point clouds without color (see Figure~\ref{fig:point_cloud}) and then evaluate FPD. Note that Dream Field~\citep{jain2021zero} does not produce 3D shapes directly, so that we cannot evaluate this work in this regard.

To further assess the text-shape consistency, we conduct a human perceptual evaluation which is detailed as follows.

\vspace*{-3pt}
\paragraph{Human perceptual evaluation Setup.}
First, we prepare generated results for human evaluation. For each input text, 
%\phil{need at least a sentence to describe how each result was produced... why nine? mention nine competitor methods? mention that the next subsection will describe the details? need to say something}
we produce nine results from state-of-the-art methods, including  CLIP-Forge~\citep{sanghi2021clip} and Dream Fields~\citep{jain2021zero}, six baseline methods, and our full method; see Section 4.2 in the main paper and Section~\ref{sec:ablation_supp:setup} in this supplementary material for details of each baseline.
%(as shown in Figure 5 in the main paper and Figure~\ref{fig:quality_supp} in this supplementary material)
Second,  we invite 10 volunteers (3 females and 7 males; aged from 19 to 58; all with normal vision) to evaluate the results. We show these results to the participants in random order without revealing how each result is produced.
%\phil{without revealing to them how each result was produced}; 
Then, they are asked to give a score from $\{1, 0.5, 0\}$ (1: perfect match, 0.5: partial match; and 0: don't match) on the degree of match between the generated shapes and input text.
%compare whether the results match the input text or not.
%Specifically, they are asked to give a score of one for perfect matches, a score of 0.5 for partial matches, and zero if not match at all. 
Then, for each method, we gather the evaluation scores from all participants and obtain the ``Consistency Score'' as $s/n$, where $s$ is the total score and $n$ is the number of samples.  
%In our evaluation, we recruited 10 volunteers (3 females and 7 males; aged from 19 to 58; all with normal vision). 

\paragraph{Details of human perceptual evaluation results.}

We show Consistency Score and the preferences in the A/B/C test from each volunteer in Tables~\ref{tab:consistent} and~\ref{tab:ABC}. As shown in Table~\ref{tab:consistent}, all ten volunteers consistently give the highest Consistency Score to our approach, and in the A/B/C test (see Table~\ref{tab:ABC}), all ten volunteers prefer results from our approach. The above further manifest the superiority of our model. %beyond the state of the arts~\citep{sanghi2021clip,jain2021zero} and all the baselines.

\begin{table}
\centering
\caption{Total scores $s$ from the ten volunteers out of 52 generated shapes.}
\label{tab:consistent}
\scalebox{0.68}{
  \begin{tabular}{ccccccccccccc}
    \toprule
     Method &  1 & 2 & 3 & 4 &5 & 6 & 7& 8& 9 & 10 & mean $\pm$ std & Consistency Score (\%) $\uparrow$ \\
    \midrule
 CLIP-Forge  & 21 &
35 &
34&
16.5&
18.5&
29&
21.5&
4&
21&
17 & 21.75 $\pm$ 9.16  & 41.83 $\pm$ 17.62 \\
     Dream Fields & 13&
17.5&
7&
19&
6.5&
22&
22&
9&
10&
6 & 13.2 $\pm$ 6.41 & 25.38 $\pm$ 12.33 \\
    \midrule
 $E_{\text{I}}$+$D$ & 14&
9.5&
5&
18&
4.5&
25&
15.5&
4.05&
4.5&
9 &10.91 $\pm$ 7.06 & 20.97 $\pm$ 13.59\\
     w/o stage-1 & 1.5&
1&
0&
1.5&
0&
3.5&
2&
0&
0&
0.5 & 1.00 $\pm$ 1.15 & 1.92 $\pm$ 2.22\\
     w/o stage-2 & 20&
14.5&
8.5&
23.5&
7.5&
27&
19&
8.5&
4.5&
20.5 & 15.35 $\pm$ 7.73 & 29.52 $\pm$ 14.86\\
     w/o $L_{\text{bg}}$ & 17.5&
15.5&
8.5&
21&
8&
27&
26.5&
9.5&
5&
22.5 & 16.1 $\pm$ 8.06 & 30.96 $\pm$ 15.49\\
    \midrule
 GLIDE+DVR & 5&
3.5&
1.5&
10&
2&
13&
6.5&
0.5&
1&
3 &4.60 $\pm$ 4.13 & 8.85 $\pm$ 7.94\\
     LAFITE+DVR & 23&
33.5&
31&
23&
27&
31.5&
27.5&
14.5&
32.5&
27.5 & 27.10 $\pm$ 5.75 & 52.12 $\pm$ 11.05\\
    \midrule
   ISS (ours) & \textbf{32}&
\textbf{37}&
\textbf{35.5}&
\textbf{29.5}&
\textbf{29}&
\textbf{37}&
\textbf{29}&
\textbf{17.5}&
\textbf{32.5}&
\textbf{33} & \textbf{31.20} $\pm$ \textbf{5.69}  & \textbf{60.00} $\pm$ \textbf{10.94}\\
    \bottomrule
  \end{tabular}
  }
\end{table}

\begin{table}
\centering
\caption{A/B/C Test results of the ten volunteers. The numbers in the table indicate the number of shapes from the corresponding method he/she likes most out of the three candidates. Volunteers can optionally select ``pass'' instead of ``A/B/C'' if he cannot decide which one is the best.}
\label{tab:ABC}
\scalebox{0.78}{
  \begin{tabular}{ccccccccccccc}
    \toprule
    Category & Method & 1 & 2 & 3 & 4 & 5 & 6 & 7 & 8 & 9 & 10 & mean $\pm$ std  $\uparrow$ \\
    \midrule
Existing works &  CLIP-Forge  & 9& 
17& 
12& 
13& 
6& 
9& 
3& 
6& 
6& 
8& 8.9 $\pm$ 4.12 \\
    \midrule
  SOTA Text2Image+SVR  & LAFITE+DVR & 9& 
16& 
9& 
12& 
7& 
13& 
9& 
20& 
8& 
14& 
11.7 $\pm$ 4.11
    \\
    \midrule
    Ours & ISS & \textbf{27}&
\textbf{19}&
\textbf{21}&
\textbf{20}&
\textbf{17}&
\textbf{25}&
\textbf{19}&
\textbf{26}&
\textbf{13}&
\textbf{30}&
\textbf{21.7} $\pm$ \textbf{5.19} \\
    \bottomrule
  \end{tabular}
  }
\end{table}

\section{Ablation Studies}\label{sec:ablation_supp}

\subsection{Baseline setups}\label{sec:ablation_supp:setup}

We create the following baselines in our ablation study. 
The first six baselines aim to assess the effectiveness of  key modules in our approach, whereas the last two adopt state-of-the-art text-to-image generation approaches to first create images then adopt DVR~\citep{niemeyer2020differentiable} to generate shapes for a fair comparison with our approach. % \phil{with our approach}.
Note that we do not adopt the most recent SVR model SS3D~\citep{alwala2022pre} (which aims to work with in-the-wild images), %\phil{(which aims to work with in-the-wild images}), 
due to its inferior generative quality and lack of texture generation.
%(lzz:Actually, one of the most recent.)}
%
\begin{itemize}[leftmargin=0.5cm]
\item $E_{\text{I}}+D$: As the first empirical study in Section 3.2 in the main paper, $E_{\text{I}}$ is adopted to extract the image feature $f_{\text{I}}$ and $D$ is trained for 3D shape generation from $f_{\text{I}}$ without the two-stage feature-space alignment.
\item w/o Stage 1: Stage-2 alignment is optimized from randomly initialized $M$, without stage 1. 
\item w/o Stage 2: Generate with $M$ after stage 1, without the test-time optimization of stage 2. 
\item w/o $L_{\text{bg\_1}}$: Remove $L_{\text{bg}}$ in stage-1 alignment.
\item w/o $L_{\text{bg\_2}}$: Remove $L_{\text{bg}}$ in stage-2 alignment.
\item w/o $L_{\text{bg}}$: Remove $L_{\text{bg}}$ in both stage-1 and stage-2 alignment.
\item GLIDE+DVR: Use a recent zero-shot text-guided image generation approach GLIDE~\citep{nichol2021glide} %\footnote{\url{https://paperswithcode.com/sota/zero-shot-text-to-image-generation-on-coco}}
to generate image $I$ from $T$, then use DVR~\citep{niemeyer2020differentiable} to generate $S$ from $I$.
\item LAFITE+DVR: Train a recent text-guided image generation approach LAFITE~\citep{zhou2021lafite} %\footnote{\url{https://paperswithcode.com/sota/text-to-image-generation-on-coco}}
with ShapeNet images, then generate image $I$ from $T$. Further generate $S$ from $I$ with DVR~\citep{niemeyer2020differentiable}.
\end{itemize}

\subsection{Quantitative and qualitative comparisons }

%\vspace*{-3pt}
%\subsubsection{Quantitative and qualitative comparisons} %\xjqi{Qualitative results are shown in Figure xxx. add a summary about your major conclusion about different approaches. For each part, you should echo the claims in the introduction to demonstrate the effectiveness of the methods.} 

In this section, we analyze the results of the above baselines one by one. %, demonstrating the superiority of ISS over the existing works and baselines. 

%Here, we only discuss (e)-(h) in Figure~\ref{fig:quality} and more details are included in the supplementary material.

%Here, we analyze the results in Figure~\ref{fig:quality_supp} (c) and (d), to compensate with the analysis shown in Table 1 and Figure 5 (e) to (i) in the main paper Section 4.3. 

\begin{figure*}
\centering
\includegraphics[width=0.99\textwidth]{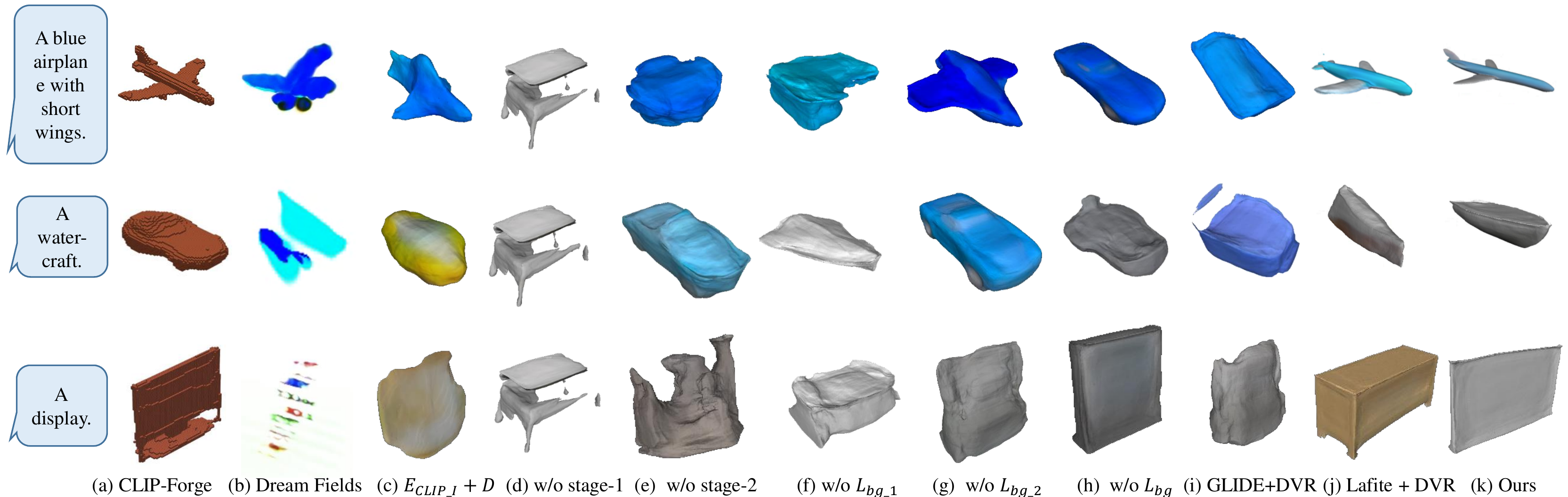}
\vspace*{-1.5mm}
\caption{Additional qualitative results compared with existing works and baselines. 
}
\label{fig:quality_supp}
\vspace*{-2.5mm}
\end{figure*}

\begin{figure*}[!t]
\centering
\includegraphics[width=0.99\textwidth]{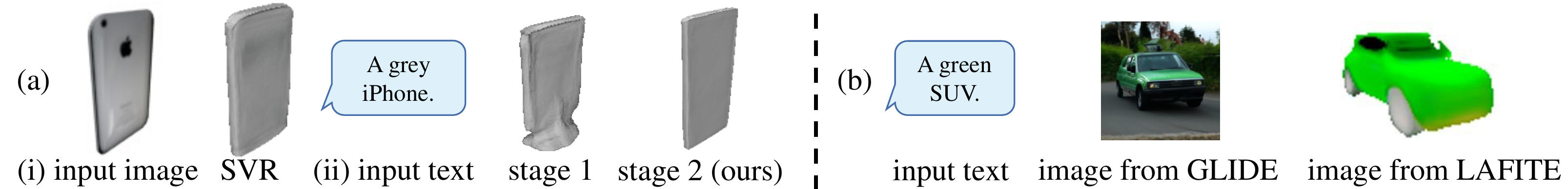}
\vspace*{-1.5mm}
\caption{A further investigation on baselines ``w/o stage 2'', ``GLIDE+DVR'', and ``LAFITE+DVR''.
%\phil{which ones... list here}.
%Can you put the sentence inside caption (a) here? Then, i can edit directly; (b) as well. Thx
%
%(a) “w/o stage 2” produces a plausible shape ("SVR") from image but a low-quality shape ("stage 1") from text; 
%further fine-tuning in Stage 2 leads to a more plausible shape from text ("stage 2").
%OR
(a) ``w/o stage 2'' produces a plausible shape (``SVR'') from image but a low-quality shape (``stage 1'') from text; 
further fine-tuning using stage-2 alignment enables us to produce a more plausible shape from text (``stage 2 (ours)'').
(b) GLIDE / LAFITE generate out-of-domain and inferior-quality images, limiting the performance of subsequent 3D generation.
%\lzz{out of domain is not good}
%\phil{:)
%
%“SVR” & “stage 1” : results of “w/o stage 2” with image and text input; “stage 2 (ours)”: after stage-2 alignment.
%
%(a) “w/o stage 2” produces a plausible shape from image, but a low-quality shape from text. “SVR” & “stage 1” : results %of “w/o stage 2” with image and text input; “stage 2 (ours)”: after stage-2 alignment.
%
%(b)  GLIDE / LAFITE can generate out-of-domain and   inferior-quality images, limiting the performance of the subsequent %3D generation. 
%
%}\lzz{(b) and-> or?}
%shown We show rendered images from two different views. }
}
\label{fig:ablation_discuss}
\vspace*{-2.5mm}
\end{figure*}

\begin{itemize}[leftmargin=0.5cm]
\item $E_{\text{I}}+D$: column (c) shows that the results generated from CLIP space $\Omega_{\text{I}}$ have inferior fidelity in terms of the texture (top row in Figure~\ref{fig:quality_supp}) and shape structure (bottom two rows in Figure~\ref{fig:quality_supp}) due to the limited capability of $E_\text{I}$ to capture details of the image. 

\item w/o Stage 1: column (d) of Figure~\ref{fig:quality_supp} indicates that without stage-1 alignment, the generated shapes are almost the same whatever text $T$ is adopted as input, since $M$ tends to map text feature $f_\text{T}$ to almost the same feature even though stage-2 alignment is enabled. It shows that a good initialization provided by stage-1 alignment is necessary for the test-time optimization of stage 2.

\item w/o Stage 2: as shown in column (e) of Figure~\ref{fig:quality_supp}, without Stage 2, may not align well with $f_S$ due to the semantic gap between $f_{\text{I}}$ and $f_{\text{T}}$.
Now, we use Figure~\ref{fig:ablation_discuss} (a) to illustrate their associated results: the model in ``w/o stage 2'' can generate reasonable shapes from a single image (see ``SVR'' in Figure~\ref{fig:ablation_discuss} (a))
%with \phil{extra?} image input\phil{s?} \lzz{a single image input, not extra}, 
but fails with text as input (see ``stage 1'' in Figure~\ref{fig:ablation_discuss} (a)); further with the stage-2 optimization, a plausible phone can be generated (see ``stage 2 (ours)'' in Figure~\ref{fig:ablation_discuss} (a)). 
%\xjqi{please also include the distance after the two steps alignment}
%
w/o $L_{\text{bg\_1}}$,  w/o $L_{\text{bg\_2}}$, w/o $L_{\text{bg}}$: stage-2 alignment cannot work properly without $L_{\text{bg}}$ in either stage-1 or stage-2 alignment or both (see column (f, g, h) of Figure~\ref{fig:quality_supp}) due to the lack of 
%awareness of the 
foreground awareness. Even though stage-1 alignment has encouraged the background to be white, we still need this loss in stage 2 to get satisfying results.

\item GLIDE+DVR: the images created by GLIDE~\citep{nichol2021glide} have a large domain gap from the training data of DVR (see Figure~\ref{fig:ablation_discuss} (b)), severely limiting the performance of GLIDE+DVR (see Figure~\ref{fig:quality_supp} (i)).

\item LAFITE+DVR: as shown in Figure~\ref{fig:quality_supp} (j), some generative results of LAFITE+DVR can be coarse (first row of the main paper and the last row of Figure~\ref{fig:quality_supp}  in this supplementary material) due to the error accumulation of the isolated two steps,~\ie, LAFITE  (see Figure~\ref{fig:ablation_discuss} (b) ``image from LAFITE'') and DVR, and some do not match the input text due to the semantic gap between $f_{\text{I}}$ and $f_{\text{T}}$ (two bottom rows of Figure 5 in the main paper and the last row of Figure~\ref{fig:quality_supp} in this supplementary material). Despite the above, subsequently generating images then shapes is still a strong baseline that is a valuable research direction in the future.

\item ours (ISS): column (k) of Figure 5 in the main paper and Figure~\ref{fig:quality_supp} in this supplementary material show that our approach can generate shapes and textures with good text-shape consistency (see ``small wings'' in the top row, ``water craft'' in the middle row of Figure~\ref{fig:quality_supp}) and fidelity, beyond all the above baselines and the existing works CLIP-Forge~\citep{sanghi2021clip} and Dream Field~\citep{jain2021zero}.

\end{itemize}

\vspace*{-3pt}
\subsection{A/B/C test} To further compare our approach with the strongest baselines CLIP-Forge~\citep{sanghi2021clip} and ``LAFITE+DVR'', we perform an A/B/C test with 10 volunteers %(see Section~\ref{sec:implementation})
to compare these two baselines with ours. Specifically, %we invited 10 volunteers and showed (\phil{XXX females and XXX males; aged XXX to XXX; all with normal vision?})
%\lzz{we show the above 10 volunteers (see Section~\ref{sec:implementation}) }
the results from the three approaches (per input text) in random order for all the $52$ texts.
Then, they were instructed to choose a most preferred one. The results in Table 1 ``A/B/C Test'' in the main paper show that our results are more preferred than others, outperforming~\citep{sanghi2021clip} by 143.8\% ($(21.70-8.90)/8.90$) and ``LAFITE+DVR'' by 85.5\% ($(21.70-11.70)/11.70$). % compared with ``LAFITE+DVR''. % and ours outperforms ``LAFITE+DVR'' by $5.6\%$ in terms of text-shape consistency. 
%\phil{actually, if this is a CHI paper, it is better to have a different set of volunteers to avoid bias, cause those people have seen the images already}

\vspace*{-3pt}
\subsection{Diversified Generation}

In addition, we evaluate the diversified generation results discussed in Section 3.3 in the main paper. Specifically, we generate additional two samples per input text, and then adopt FID~\citep{heusel2017gans}, FPD~\citep{liu2022towards} (the lower, the better) for the fidelity evaluation and Point Score (PS)~\citep{liu2022towards} (the higher, the better) for the diversity evaluation. The results are: \textbf{FID: 113.98}, \textbf{FPD: 35.37}, \textbf{PS: 3.11}, which is even better than our one-text-one-shape generative results (FID: 124.42 $\pm$ 5.11, FPD: 35.67 $\pm$ 1.0, PS: 3.18 $\pm$ 0.11), manifesting the superior performance of diversified generation capability of ISS. %It can be partially explained that the diversified results share a more similar distribution to the ground truth datasets for above metrics, which also contain diversified images/shapes. 
%The result manifests the superior diversified generation capability of ISS. 

\section{Discussions on Text-Guided Shape Stylization}\label{sec:stylization}

\begin{figure*}
\centering
\includegraphics[width=0.99\textwidth]{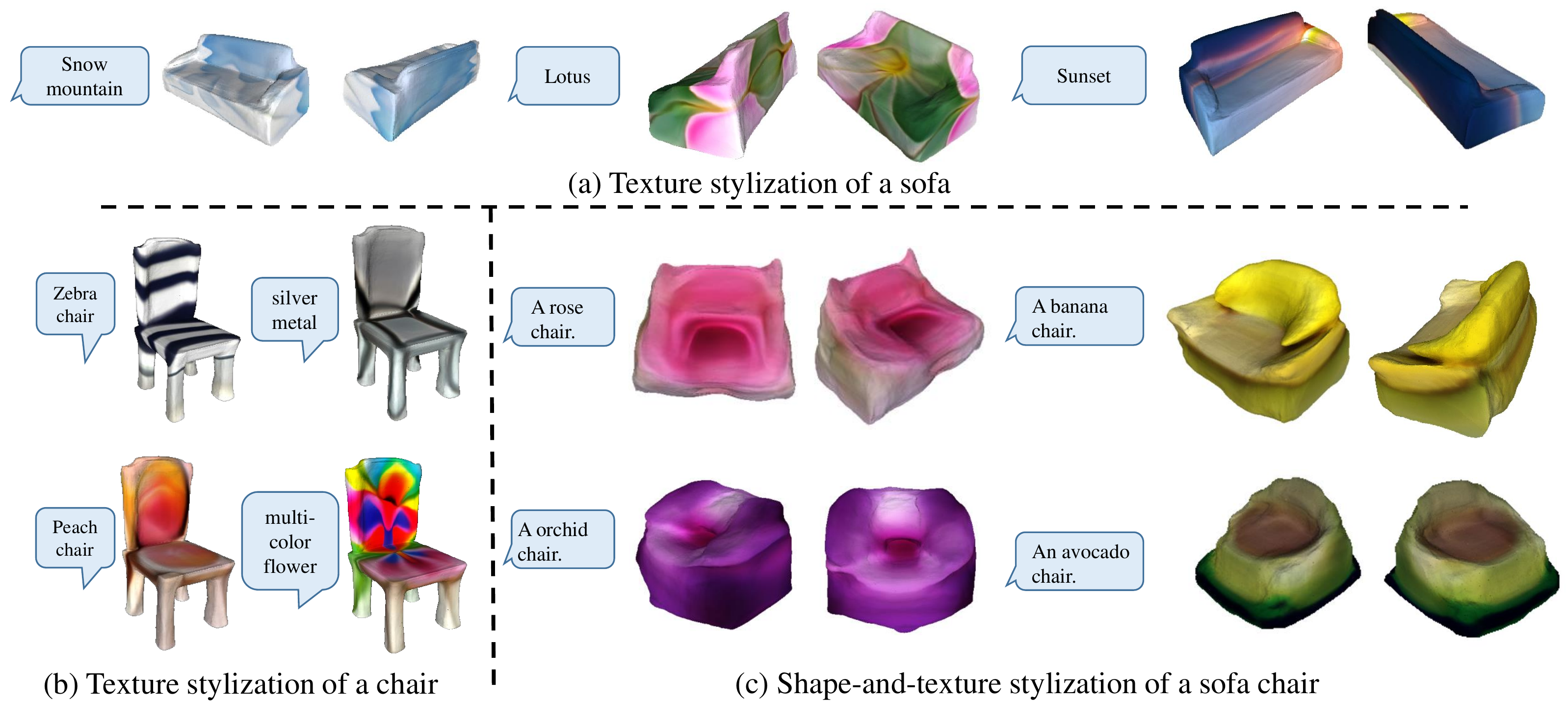}
\caption{Additional stylization results. } 
\label{fig:style_supp}
\end{figure*}

\paragraph{Additional stylization results.}
As shown in Figure~\ref{fig:style_supp} (a), our text-guided shape stylization is able to generate vivid landscape and flower textures on the sofa shape. Note that the sofa is generated from the text ``A sofa with black backrest'' (see Figure~\ref{fig:shapenet_supp} top right) with totally different initial color from the stylization results. Further, in Figure~\ref{fig:style_supp} (b), we show four additional texture stylization results in addition to the Figure 9 in the main paper. In addition, as shown in Figure~\ref{fig:style_supp} (c), our shape-and-texture stylization is able to generate novel shapes and textures beyond the dataset, and create imaginary shapes with diversified structures. Note that our results achieve a good balance on the stylization and the functionality. For example, our result of ``avocado chair'' possesses both the style of ``avocado'' and the functionality of ``chair''. % compared with the result from Dream Field~\citep{jain2021zero} with the same input text, which lacks the functionality of the chair. 

\begin{figure*}
\centering
\includegraphics[width=0.99\textwidth]{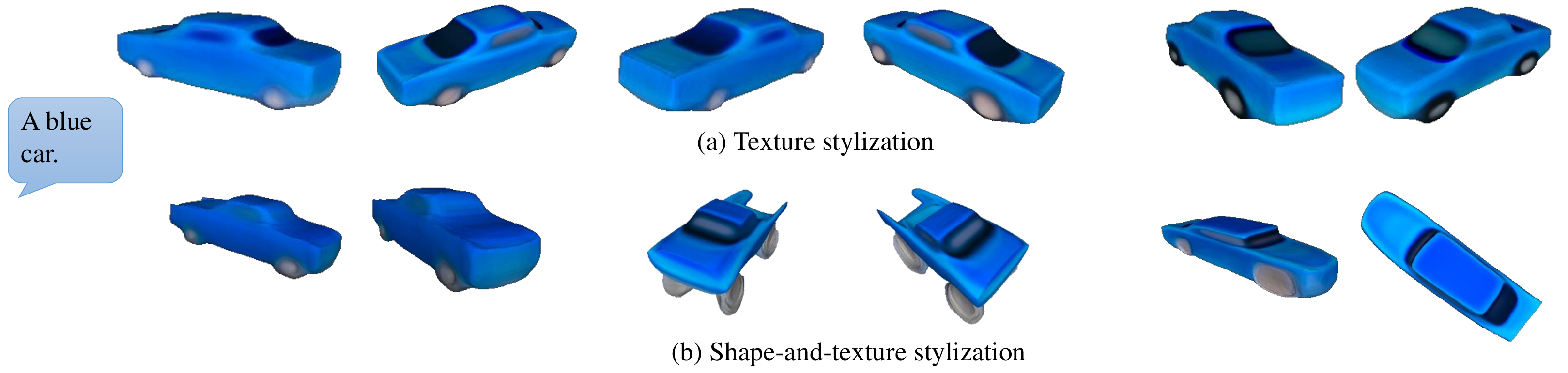}
\vspace*{-1.5mm}
\caption{Results of shape-and-texture stylization with the same text ``A blue car'' and different 3D prior loss $L_P$. }
\label{fig:blue_car}
\vspace*{-2.5mm}
\end{figure*}

As shown in Figure~\ref{fig:blue_car}, shape-and-texture stylization consistently produces the cars that is consistent with the input text ``a blue car'', with proper variation in terms of the color and shape. In addition, the degree of the shape variations can be controlled by the loss weight $\lambda_P$ of the 3D prior loss $L_p$.

\paragraph{Why does shape-and-texture stylization need an initial shape? }
Shape-and-texture stylization is initialized by the two-stage feature-space alignment result, since it provides the 3D prior.

\paragraph{Relationship of texture stylization and shape-and-texture stylization.}
Texture stylization and shape-and-texture stylization have their own merits. Texture stylization keeps the shape unchanged and is able to guarantee the functionality of the shape. In addition, it can take some abstract text descriptions as input like ``sunset'' in Figure~\ref{fig:style_supp} (a). Shape-and-texture stylization is able to generate novel and imaginary structures beyond the training dataset. However, there is a tradeoff between the stylization and the functionality.

\section{Additional Results}~\label{sec:results}

\begin{figure*}
\centering
\includegraphics[width=0.99\textwidth]{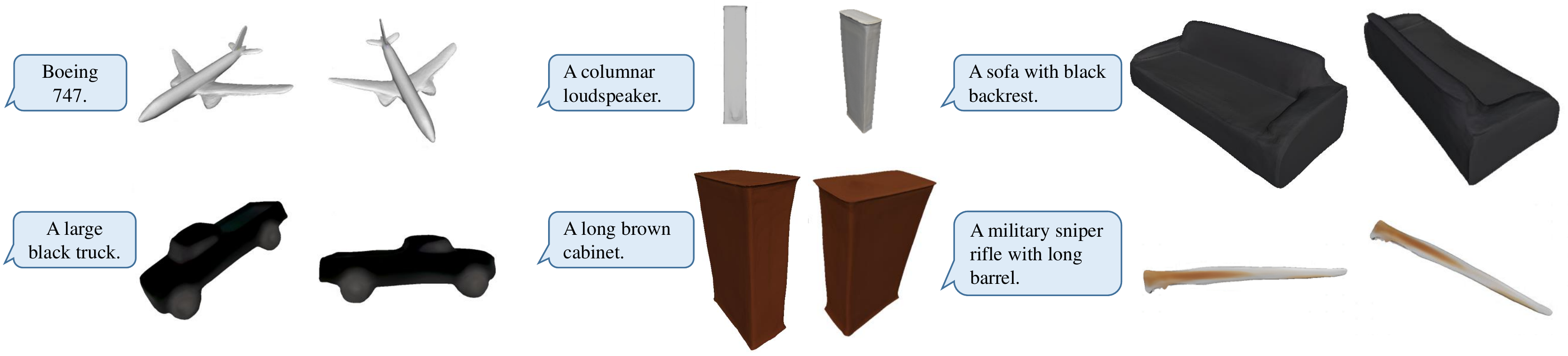}
\vspace*{-1.5mm}
\caption{Additional generative results on the ShapeNet~\citep{shapenet2015} dataset. 
}
\label{fig:shapenet_supp}
\vspace*{-2.5mm}
\end{figure*}

\paragraph{ShapeNet.}
As shown in Figure~\ref{fig:shapenet_supp}, our approach can generate view-consistent 3D shapes on ShapeNet~\citep{shapenet2015} that well match the input texts.

\begin{figure*}
\centering
\includegraphics[width=0.99\textwidth]{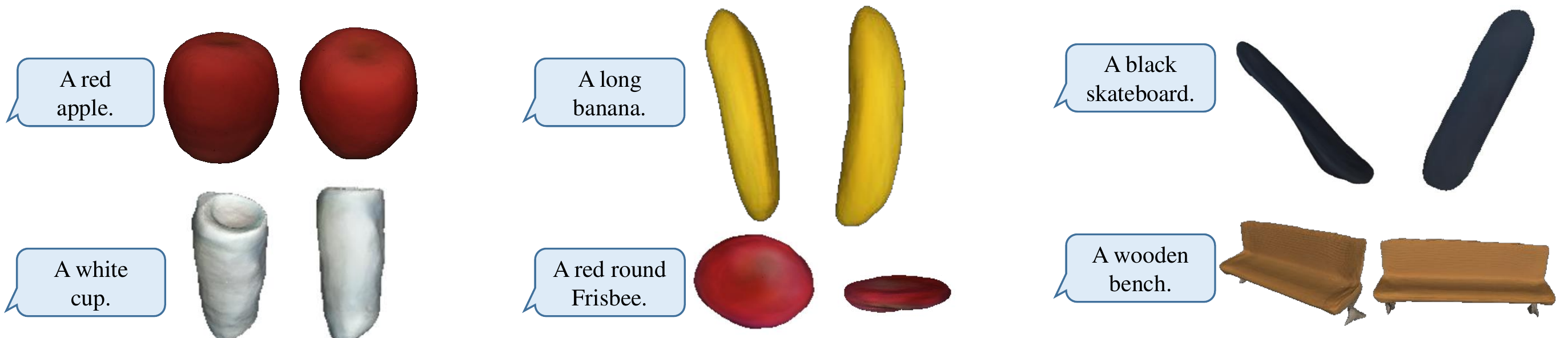}
\vspace*{-1.5mm}
\caption{Additional generative results on the CO3D~\citep{reizenstein2021common} dataset. 
}
\label{fig:co3d_supp}
\vspace*{-2.5mm}
\end{figure*}

\paragraph{CO3D.}
Further, we show more text-guided shape generation results on the CO3D~\citep{reizenstein2021common} dataset in Figure~\ref{fig:co3d_supp}. These results again manifest the capability of our approach on real-world 3D shape generation, beyond the existing works~\citep{chen2018text2shape,sanghi2021clip,liu2022towards} that focus only on the synthetic shape generation on ShapeNet~\citep{shapenet2015}.

\paragraph{Working with GET3D}

\begin{figure*}
\centering
\includegraphics[width=0.99\textwidth]{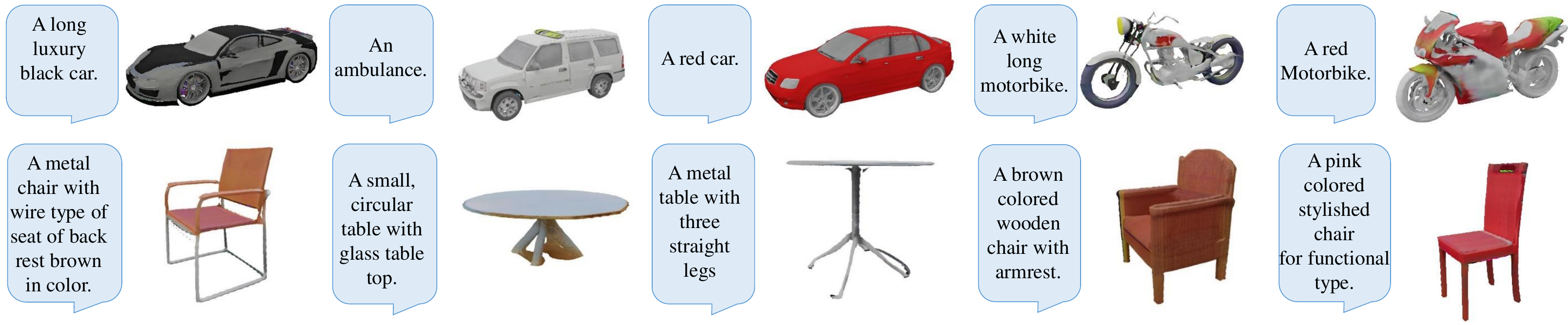}
\vspace*{-1.5mm}
\caption{Additional generative results built upon GET3D~\citep{gao2022get3d}. 
}
\label{fig:get3d}
\vspace*{-2.5mm}
\end{figure*}

When working with GET3D~\citep{gao2022get3d}, our approach can generate 3D shapes of good fidelity from texts, as shown in Figure~\ref{fig:get3d} in this supplementary file. %Note that GET3D supports text-guided 3D shape stylization to stylize a given shape, but does not support text-guided 3D shape generation, and our method can enable GET3D to do this task. 

\begin{figure*}
\centering
\includegraphics[width=0.99\textwidth]{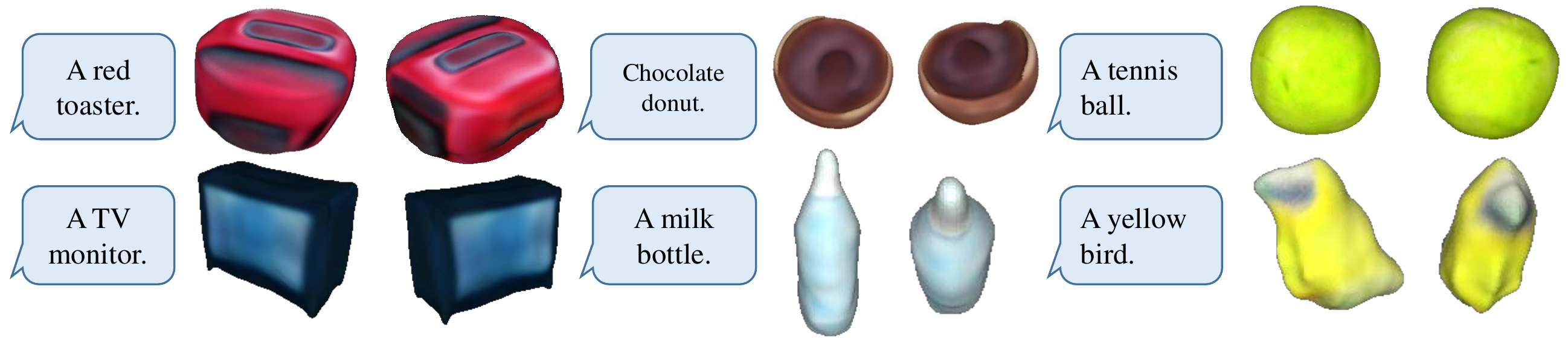}
\vspace*{-1.5mm}
\caption{Additional generative results built upon SS3D~\citep{alwala2022pre} using single images in training without camera pose. 
}
\label{fig:ss3d_supp}
\vspace*{-2.5mm}
\end{figure*}

\paragraph{Single-image categories.}
In Figure~\ref{fig:ss3d_supp}, we present more text-guided generations for more categories using single images in training without camera pose, built upon~\citet{alwala2022pre}. The results further demonstrate the compatibility of our approach to various SVR approaches, particularly generating plausible 3D shapes from text with single images in training.

\begin{figure*}
\centering
\includegraphics[width=0.99\textwidth]{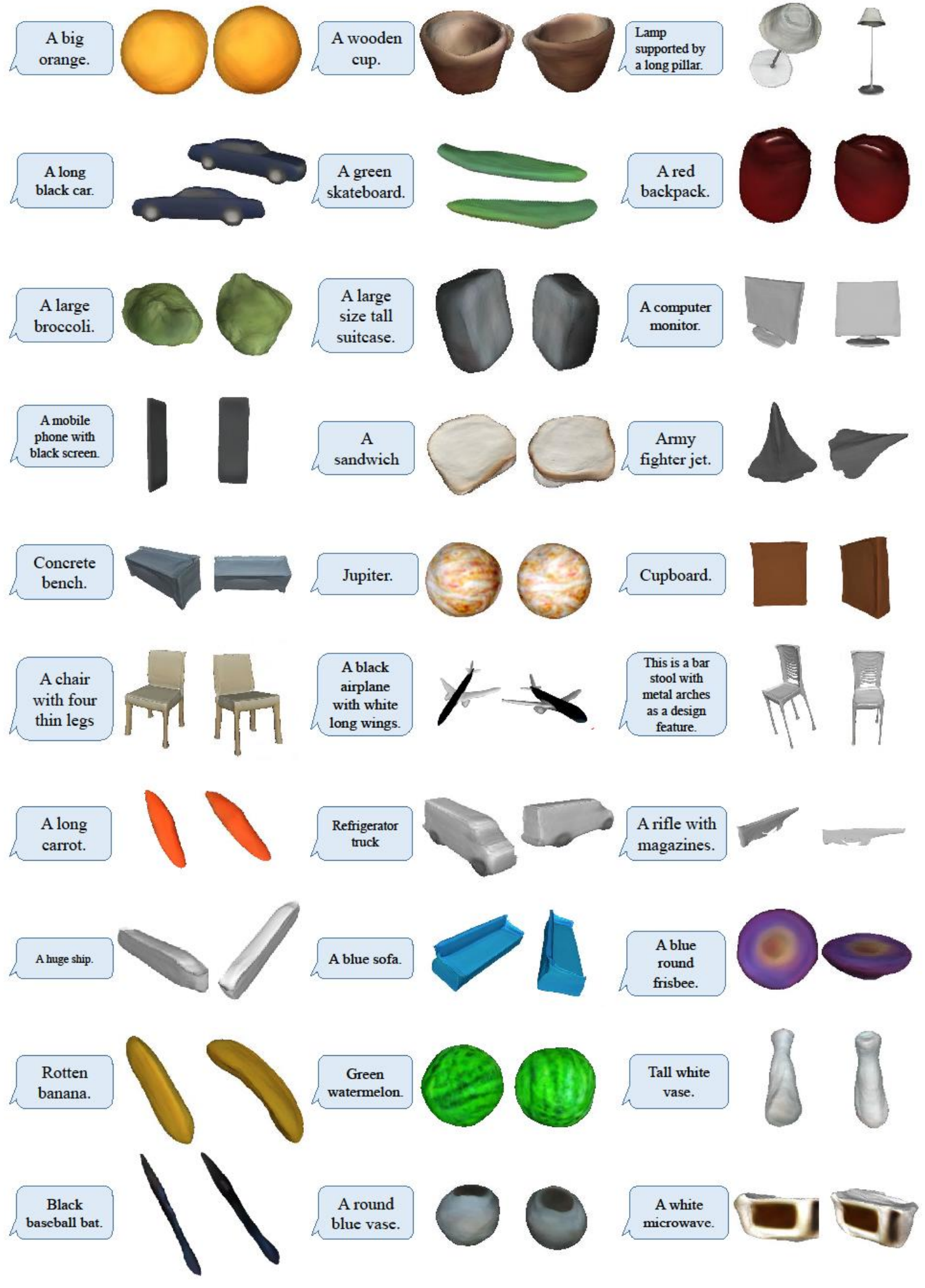}
\vspace*{-1.5mm}
\caption{Generative results of ISS. Using our new approach, we are able to effectively produce 3D shapes for a wide variety of categories from texts. 
}
\label{fig:all}
\vspace*{-2.5mm}
\end{figure*}

\paragraph{More generative results.}
Further, we present more generative results of our approach in Figure~\ref{fig:all}. Using ISS, we are able to effectively generate a wide variety of 3D shapes from texts.

\begin{figure*}
\centering
\includegraphics[width=0.9\textwidth]{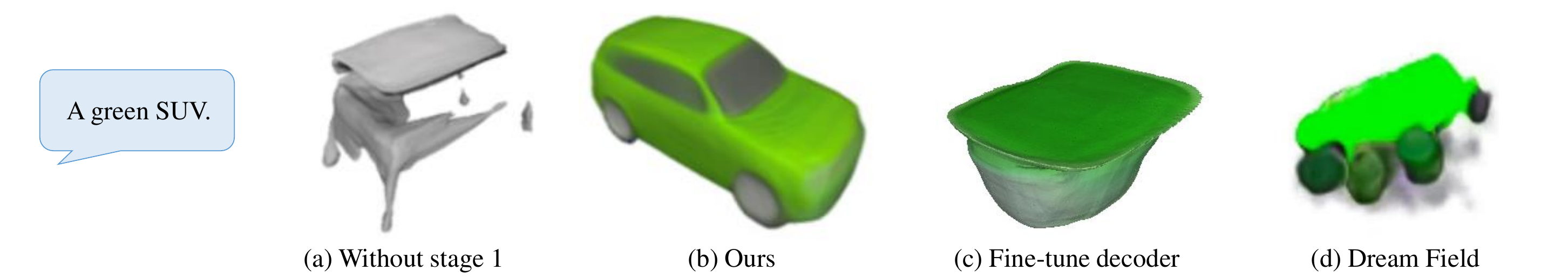}
\vspace*{-1.5mm}
\caption{Results of optimizing the decoder, instead of the two-stage feature-space alignment. }
\label{fig:suv}
\vspace*{-2.5mm}
\end{figure*}

\begin{figure*}
\centering
\includegraphics[width=0.99\textwidth]{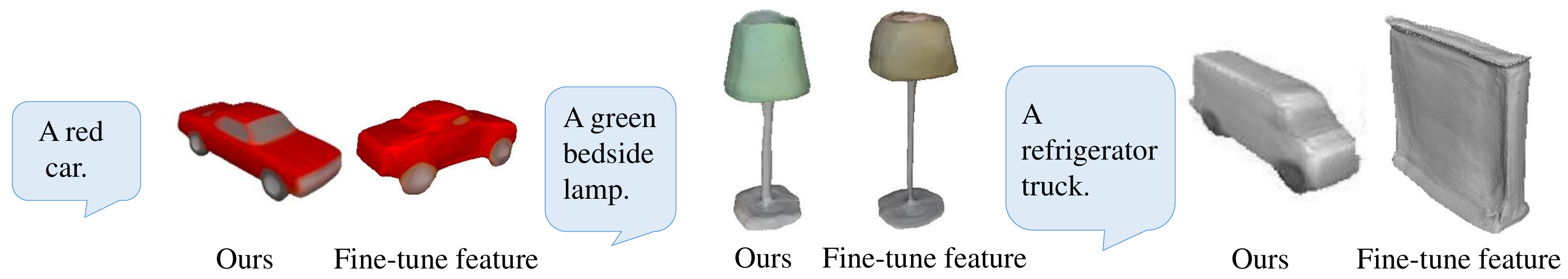}
\vspace*{-1.5mm}
\caption{Results of optimizing the feature in shape space, instead of the stage-2 feature-space alignment. }
\label{fig:feature}
\vspace*{-2.5mm}
\end{figure*}

\section{Discussions on Two Alternative Training Strategies}\label{sec:alternative}
\paragraph{Fine-tune decoder.}

In this section, we discuss an alternative training strategy to optimize the decoder directly with the CLIP consistency loss, instead of using two-stage feature space alignment.

First, we provide the results of the above training strategy. It produces unsatisfactory 3D shapes such as the one shown in Figure~\ref{fig:suv} (c) with more than 20 minutes of training. Besides, Dream Field~\citep{jain2021zero} also utilizes the same idea to optimize the decoder directly; yet, the generated shape (see Figure~\ref{fig:suv} (d)) is also far from satisfactory even though Dream Fields is trained only to produce multi-view images with an NeRF-like architecture, unlike ours that is capable for 3D shape generation.  %. Note that DF can only generate multi-view images but not 3D shapes as our method.

Then we try to analyze why the above strategy fails to produce desired shapes. Leveraging the pre-trained decoder that has already incorporated the 3D shape prior, our approach is able to search for the desired shape in the shape feature space $\Omega_S$ efficiently. %Specifically, our approach only needs 85 seconds in the process to generate a shape. 
On the contrary, fine-tuning the decoder in stage 2 without the explicit 2D/3D supervision will destroy the pre-trained shape feature space $\Omega_S$, which is used to introduce 3D priors. Specifically, in stage 2 (``alignment stage''), the model is trained at test time with the user-provided text without any explicit 2D/3D supervision, so it is hard for the model to gain the knowledge about what the desired 3D shape should be like. In other words, it is extremely challenging for the model to learn the 3D shape prior with only CLIP consistency supervision from text.

Further, we analyze why the decoder can be fine-tuned in shape-and-texture stylization. First, note that 3D shape generation is a very different task from shape-and-texture stylization, since shape-and-texture stylization is initialized by the two-stage feature-space alignment result, which has already learned the 3D prior. Therefore, we can fine-tune the decoder in shape-and-texture stylization.

\paragraph{Update the feature in the shape space.}
In this section, we discuss another alternative training strategy to optimize the feature in shape space $\Omega_S$, instead of optimizing the mapper $M$ with stage-2 feature space alignment.

%First, we show the generative results of this strategy. As shown in Figure~\ref{fig:feature}, in most cases, directly optimizing the 3D shape feature cannot generate satisfying results. It is because 

However, the shape feature has only 256 dimensions, which is far smaller than the number of weights in the mapping network.  Hence, directly optimizing the shape feature in $\Omega_S$ is not capable for generating the desired 3D shape as shown in Figure~\ref{fig:feature}, and the stage-2 feature space alignment is necessary in our approach.

\section{Analysis on Feature Space Mapping}\label{sec:mapping}

To  provide more insights on explaining how the latent space is mapped, we measure the distance of features at different stages on all the samples in our test set on ShapeNet in Table~\ref{tab:mapping}. The notations follow Figure 3 (c) of the main paper, $M$ means the mapper, and $d$ means cosine distance. Our ultimate goal is to obtain a text mapper $M'$ (Figure 2 in the main paper) to map the text feature space $f_T$ to shape feature space $f_S$ using image with features $f_I$ as a stepping stone to gradually narrow their distances using two stage mapping. Note that the image $f_I$ and text features $f_T$ are obtained using pre-trained CLIP models.  

In the stage-1 alignment process, we train a mapper $M$ to map image features $f_I$ to a space $M(f_I)$ close to the shape space $f_S$ using image data and the regression loss $L_M$. Note that the text feature $f_T$ and image feature $f_I$ are all from the CLIP model in a shared embedding space. It’s natural that the trained mapper can be used to map the text feature $f_T$ to $M(f_T)$, making text features closer to the shape space. However, when measuring the distance among $M(f_T)$,  $M(f_S)$, we find the average distance among all samples is $d(M(f_I), M(f_T))= 0.58 \pm 0.23$,  the average distance between $M(f_I)$ and $f_S$ is $0.21 \pm 0.10$ , and the average distance between shape and text is  $d(M(f_T), f_S)) = 0.45  \pm  0.20$. This implies that there is a gap between CLIP image and text feature after the first step mapper and further motivates our stage-2 alignment. Note that there is no GT shape for our task on 3D generation, so we manually select a shape in the ShapeNet dataset that matches the input text as the GT.

In the stage-2 alignment, $M$ is further updated and the final delivered mapper is called $M'$, which is to further narrow down the gap between mapped text features and shape features. The average distance between the mapped text feature $M’(f_T)$ and shape feature space $f_S$: $d(M’(f_T), f_S)) = 0.17 \pm  0.08$ which is much smaller than the corresponding distance after stage-1 alignment.  It shows that stage 2 alignment can significantly reduce the difference between the mapped text and the GT shape feature from 0.45 to 0.17 on average.

\begin{table}
\centering
\caption{Distance changes in the feature space mapping of all the 52 samples in the test set. $d$ means cosine distance. Almost all distances are consistently reduced after our stage-2 alignment.  }
\label{tab:mapping}
\scalebox{0.6}{
  \begin{tabular}{cccccc}
    \toprule
    Text & $d(M(f_I),M(f_T))$&$d(M(f_I), f_S))$&$d(M(f_T), f_S))$&$d(M’(f_T), f_S))$&$d(M(f_T), M’(f_T))$  \\
    \midrule
a glass single leg circular table &0.63&0.31&0.58&0.14&0.34\\
a wooden double layers table & 0.72&0.10&0.64&0.14&0.65 \\
a square metal table&0.76&0.32&0.44&0.21&0.30\\
a round shaped single legged wooden table&0.47&0.24&0.21&0.21&0.30\\
this is a bar stool with metal arches as a design feature&0.43&0.33&0.35&0.18&0.17\\
a children chair with little legs&0.79&0.20&0.43&0.12&0.35\\
a swivel chair with wheels&0.61&0.30&0.38&0.22&0.11\\
a red recliner seems confortable&0.33&0.20&0.23&0.24&0.05\\
a red car&1.02&0.19&0.78&0.10&0.69\\
a green SUV&0.49&0.12&0.24&0.19&0.25\\
a large black truck&0.74&0.19&0.50&0.15&0.53\\
a long luxury black car&0.67&0.22&0.54&0.09&0.54\\
army fighter jet&0.43&0.54&0.43&0.25&0.28\\
a black airplane with long white wings&0.62&0.12&0.32&0.16&0.10\\
a blue airplane with short wings&0.51&0.14&0.55&0.33&0.40\\
boeing 747& 0.35&0.13&0.23&0.06&0.20\\
a big ship for transportation&0.95&0.17&0.72&0.35&0.21\\
a boat with sail&0.39&0.19&0.46&0.12&0.45\\
a watercraft&0.12&0.14&0.27&0.22&0.06\\
a wooden boat&0.76&0.29&0.52&0.12&0.49\\
a blue sofa&0.46&0.20&0.34&0.12&0.27\\
sofa with legs&0.67&0.13&0.55&0.10&0.35\\
a sofa with black backrest&0.59&0.14&0.33&0.08&0.16\\
a small sofa&0.49&0.20&0.33&0.10&0.28\\
a long brown bench&0.57&0.29&0.34&0.18&0.27\\
a marble bench&0.61&0.16&0.46&0.12&0.29\\
a metal bench&0.18&0.29&0.20&0.20&0.09\\
concrete bench&0.48&0.07&0.46&0.11&0.36\\
a military sniper rifle with long barrel&0.47&0.14&0.29&0.07&0.21\\
a rifle with magazines&0.71&0.48&0.51&0.17&0.43\\
a short rifle&0.75&0.15&0.81&0.35&0.52\\
rifle shotgun&0.38&0.13&0.25&0.10&0.09\\
a computer monitor&0.28&0.07&0.22&0.07&0.20\\
a display&0.22&0.18&0.15&0.13&0.02\\
a monitor with square base&0.56&0.17&0.75&0.33&0.34\\
a TV monitor&0.70&0.16&0.41&0.11&0.27\\
a cabinet with cylindrical legs&0.53&0.24&0.27&0.18&0.11\\
a cupboard&0.54&0.37&0.42&0.19&0.45\\
a long brown cabinet&0.57&0.29&0.34&0.18&0.27\\
a wardrobe&0.41&0.13&0.32&0.16&0.26\\
a desk lamp&0.92&0.35&0.51&0.16&0.45\\
bedside lamp&0.71&0.48&0.55&0.17&0.43\\
lamp supported by a long pillar&0.19&0.12&0.18&0.05&0.14\\
mushroom-like lamp&1.20&0.17&1.06&0.32&0.46\\
a mobile phone&0.75&0.09&0.85&0.28&0.55\\
a small cell phone&0.49&0.20&0.33&0.10&0.28\\
a mobile phone with black screen&0.94&0.39&0.61&0.14&0.68\\
an iphone&0.75&0.25&0.54&0.22&0.57\\
a columnar loudspeaker&0.66&0.25&0.54&0.15&0.40\\
a loudspeaker with metal surface&0.34&0.17&0.44&0.16&0.30\\
a wooden loudspeaker&0.72&0.10&0.64&0.14&0.65\\
a cylindrical loudspeaker&1.06&0.19&0.82&0.31&0.27\\
    \midrule
\textbf{mean $\pm$ std}&\textbf{0.58 $\pm$ 0.234}&
\textbf{0.21 $\pm$  0.10}&
 \textbf{0.45  $\pm$  0.20} &
 \textbf{0.17 $\pm$ 0.08}&
 \textbf{0.32 $\pm$ 0.17}\\
    \bottomrule
  \end{tabular}
  }
\end{table}

\section{Related Works on Single-View Reconstruction and Differentiable Rendering}\label{sec:review}

\paragraph{Single-view reconstruction (SVR)} This task aims to reconstruct the 3D shape from a single-view image of it. Recently, many approaches have been proposed for meshes~\citep{agarwal2020gamesh}, voxels~\citep{zubic2021effective}, and 3D shapes~\citep{niemeyer2020differentiable}. Recently, to extend SVR to in-the-wild categories,~\citet{alwala2022pre} proposed a new approach called SS3D to learn 3D shape reconstruction using single-view images for the reconstruction of hundreds of categories.

As 3D-to-2D projections, single-view 2D images are more closely related to shapes than texts, since they reveal many attributes of 3D shapes, e.g., structure, details, appearance, etc.
The strong correlation between 2D images and 3D shapes motivates us to reduce the challenging text-to-shape generation task to text-to-image and then SVR, by connecting the CLIP features from text with the shape features in SVR using images as an intermediate step to gradually bridge the gap between text and shape.
Specifically, we extend the pre-trained SVR model to be compatible with text input, transforming the challenging text-to-shape task into SVR. Our framework can work with different SVR approaches to extend them for 3D shape generation from texts. So, our approach is orthogonal to the SVR approaches.

\paragraph{Differentiable Rendering} 
3D rendering is an important topic in computer vision and graphics. It takes a 3D scene as input and predicts the 2D view of it given a camera pose. Beyond 3D rendering, differentiable rendering further aims to derive the differentiations of the rendering function. With differentiable rendering, a renderer can be integrated into an optimization framework, thus a 3D shape can be reconstructed from multi-view 2D images. 
Neural Volume Rendering~\citep{mildenhall2020nerf} and its following works~\citep{jain2021putting,barron2021mip} aim to synthesize novel view images of a 3D scene. Besides, recent works ~\citep{niemeyer2020differentiable,munkberg2022extracting,gao2022get3d} leverage differentiable rendering for 3D shape generation using 2D images. In this work, we derive 2D images of the generated 3D shape using differentiable rendering and use a pre-trained large-scale image-language model CLIP to encourage the 2D images to be consistent with the input text. Thanks to differentiable rendering, we can update the generated 3D shape indirectly using the rendered images.

\section{Analysis on the Number of Parameters}~\label{sec:parameter}

%\textcolor{blue}{In this section, we compare the number of parameters of CLIP-Forge~\cite{sanghi2021clip}, Dream Fields~\cite{jain2021zero} and our approach in Table~\ref{tab:parameter}. }
Our model incorporates only a small number of learnable parameters for shape generation in $M$. Note that the parameters of the SVR model are not tuned for the specific text-to-shape generation task. Therefore, they are not counted in the total number of parameters.

We provide the comparison of learnable model parameters responsible for shape generation and compare it with existing methods in Table~\ref{tab:parameter}. 

\begin{table}
\centering
\caption{Number of parameters and performance of existing works and our method. }
\label{tab:parameter}
\scalebox{0.75}{
  \begin{tabular}{ccccc}
    \toprule
   Method &Number of parameters (M)& FID & FPD & Consistency Score \\
    CLIP-Forge~\citep{sanghi2021clip}& Normalized flow network: 18.37& 
162.87& 
37.43& 
41.83±17.62\\
    Dream Fields~\citep{jain2021zero} &0.61&181.25&N.A.&25.38±12.33\\
        \midrule
    Ours& Mapper: 2.43&
\textbf{124.42±5.11}&
35.67±1.09&
60.0±10.94\\
    Ours with a lightweight mapper & \textbf{Mapper: 0.46}&129.01&\textbf{34.26}&\textbf{67.21±10.64} \\

    \bottomrule
  \end{tabular}
  }
\end{table}

%\textcolor{blue}{Our mapper has only 2.43M parameters, vs. 18.37M parameters in CLIP-Forge~\cite{sanghi2021clip} in their normalizing flow network; their model is more than 8 times in size, compared with our model. Note that we do not include the parameters of CLIP-Forge auto-encoder for a fair comparison. With less parameters, our model outperforms CLIP-Forge in terms of FID, FPD and Consistency Score consistently as shown in Table 1 in the main paper.}

%\textcolor{blue}{Dream Field~\cite{jain2021zero} has only 0.61M parameters, which is smaller than ours. To show that our performance gain is not from more learnable parameters, we design a lightweight mapper composed of only three MLP layers with 512, 256, 256 output neurons respectively. This mapper contains only 0.46 M parameters (512$\times$512+512$\times$256+256$\times$256) in total, which is even less than Dream Fields. Our new lightweight mapper achieves performance FID: 129.0, Consistency Score: 56.15±15.63, which still outperforms Dream Fields (FID: 181.25, Consistency Score: 25.38 ± 12.33) by a large margin. Please see Figure~\ref{fig:lightweight} in this supplement material for the associated visualizations. }

%\textcolor{blue}{Also, we would like to clarify why we can achieve a better performance with fewer parameters. It is because our approach is built upon a pre-trained SVR model.  We train a small mapper network to work with a pre-trained SVR model and leverage a novel two-stage feature space alignment approach to achieve good performance.}

The number of learnable parameters of CLIP-forge~\citep{sanghi2021clip} is 8 times larger than ours. Note that we do not include the parameters of the CLIP-Forge auto-encoder for a fair comparison. With much fewer parameters, our model outperforms CLIP-Forge in all evaluated metrics (see Table~\ref{tab:parameter}). This demonstrates that our performance gain is not purely from the learnable parameters. 

\begin{figure*}
\centering
\includegraphics[width=0.99\textwidth]{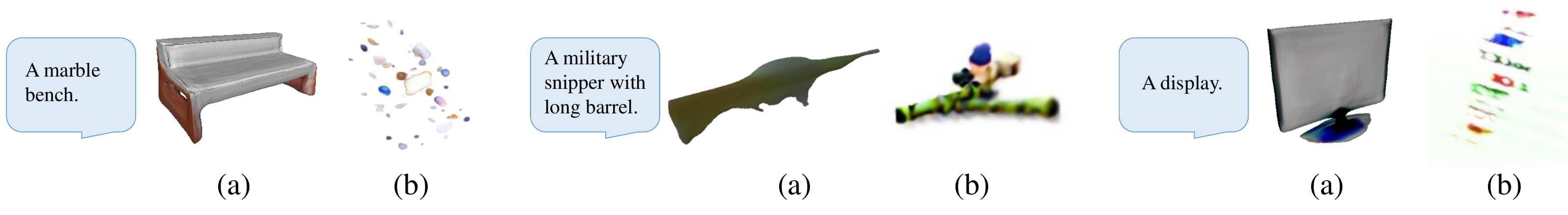}
\vspace*{-1.5mm}
\caption{(a) Our generative result with lightweight mapper vs. (b) Dream Fields~\citep{jain2021zero}. 
}
\label{fig:lightweight}
\vspace*{-2.5mm}
\end{figure*}

To better compare with Dream Fields~\citep{jain2021zero} (0.61 M parameters), we design a lightweight mapper to match the total number of parameters. Specifically, the mapper is composed of three fully connected layers with 512, 256, and 256 output dimensions, yielding a total of 0.46M parameters which is smaller than Dream Field. With this new mapper, we still outperform Dream Fields in terms of metrics. Please see Figure~\ref{fig:lightweight} for quantitative comparison. This further demonstrates our major performance gain is not from the learnable parameters. 

From the results in Table~\ref{fig:lightweight}, we can consistently observe that our model can achieve much better performance with much fewer parameters, manifesting the efficiency of our proposed image as a stepping-stone pipeline that allows us to leverage the 3D priors in pre-trained SVR models to enable text to 3D shape synthesis without requiring paired text and 3D data. 

Note that we exclude the number of parameters of the decoder $D$ in our comparison because the only effect of fine-tuning $D$ is to make the background white and does not contribute to the generative capability of our model. As shown in Figure 2 of the main paper, mapper $M$ and decoder $D$ are trained with their own losses \textbf{separately} at the same time by stopping the gradients from $L_D$ and $L_{bg}$ to propagate to $M$.

Also, to show that fine-tuning $D$ does not improve its generative capability, we feed the same input feature $f_s$ to $D$ before and after fine-tuning, they generate almost the exact same 3D shape as the original SVR model, as shown in Figure~\ref{fig:same} in the supplementary material. And in stage 2, $D$ is not optimized. Admittedly, our generative capability benefits from the SVR model including $D$. However, $D$ is not optimized for our text-to-shape generation task. Therefore, we exclude the number of parameters of decoder $D$ for fair comparison.

%\textcolor{blue}{\paragraph{Why we only count the parameters in the mapper $M$ but not decoder $D$.} It is because mapper $M$ and decoder $D$ are trained at the same time but \textbf{separately}. As shown in Figure 2 of the main paper, we finetune the SVR decoder $D$ with the original SVR loss $L_D$ (to preserve the 3D shape generation capability) and an additional background loss $L_{bg}$ (to encourage the rendered image to have white background). At the same time, we train mapper $M$ with regression loss $L_M$. To summarize, the decoder $D$ and mapper $M$ are trained at the same time but \textbf{separately}, and we achieve this by \textbf{stopping the gradient} between $M$ and $D$ to avoid the effects of $L_D$ and $L_{bg}$ on $M$. Hence, the only effect of finetuning of $D$ is to make the background white, without affecting the generative capability. To show this point, we feed the same input feature $f_s$ to $D$ before and after fine tuning, they generate almost the exact same 3D shape as the original SVR model, as shown in Figure~\ref{fig:same} in this supplement file. In other words, we can optionally first fine-tuning $D$ using $L_D+L_{bg}$ as a post-processing of the SVR model, which is a one-time pre-training. Then we train our model, i.e., train $M$ with regression loss $L_M$, without updating $D$.}

%\textcolor{blue}{And in stage-2 alignment, $D$ is frozen. Therefore, we can only count the parameters in $M$ for comparison with CLIP-Forge and Dream Fields. }

\begin{figure*}
\centering
\includegraphics[width=0.99\textwidth]{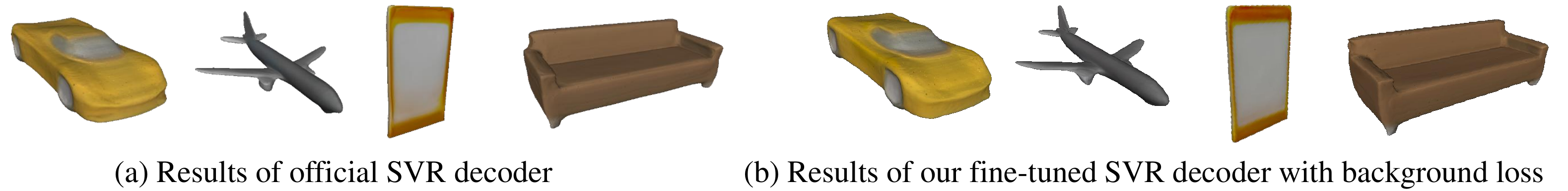}
\vspace*{-1.5mm}
\caption{The generative results of the original SVR model and after our stage-1 feature space alignment using the same feature $f_S$. 
}
\label{fig:same}
\vspace*{-2.5mm}
\end{figure*}

\section{Failure Cases}~\label{sec:failure}

Here are some examples of failure cases of our approach.

\paragraph{The complex and unusual shapes, \eg, ``an oval table with 3 legs. ''}

\begin{figure*}
\centering
\includegraphics[width=0.99\textwidth]{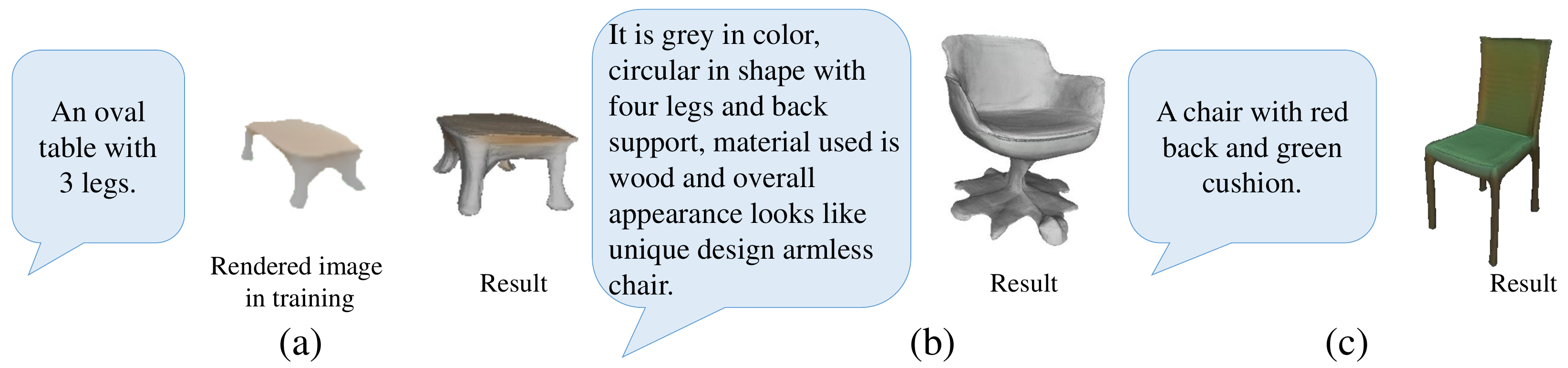}
\vspace*{-1.5mm}
\caption{Failure cases of our approach. 
}
\label{fig:failure}
\vspace*{-2.5mm}
\end{figure*}

First, our model fails to generate the shape from the text ``an oval table with 3 legs. ''. Our approach leverages the CLIP consistency loss in the rendered 2D image; however, in the rendered image shown in Figure~\ref{fig:failure} (a), only three legs can be seen and the remaining one is occluded, which confuses the model training.

\paragraph{The given text is very long, \eg, ``it is grey in color, circular in shape with four legs and back support, material used is wood and overall appearance looks like unique design armless chair. ''}

Our model fails to generate the shape from the above long description as shown in Figure~\ref{fig:failure} (b). Some attributes are missing including ``armless'', ``four legs''. This is partially due to the limited representative capability of a single CLIP feature for such a long sentence. We may incorporate an additional local feature like~\citet{liu2022towards} to handle the long text in the future.  
%Due the limit of the GPU memory, the rendered image in training has the limited resolution (100$\times$100 in our experiments. It makes our model hard to generate tiny structures since they cannot rendered clearly in training. }

\paragraph{The shapes with multiple fine-grained descriptions, \eg, ``A chair with a red back and a green cushion.''}

As shown in Figure~\ref{fig:failure} (c), our model fails to generate the shape ``A chair with a red back and a green cushion. '' As studied in some recent works~\citep{yao2021filip,li2022fine}, CLIP mainly address on the global image feature, but has inferior capability to capture fine-grained features. Therefore, our approach may fail to generate shapes where multiple fine-grained descriptions are given. In the future, we may try more recent pre-trained text-and-image embedding models to enhance the model's capability to handle the fine-grained descriptions.

\section{Limitations}~\label{sec:limitation}

This work still has some limitations. First, our performance is limited by the SVR model that our approach is built upon, \eg, some results in Figure 10 of the main paper and Figure~\ref{fig:ss3d_supp} in this supplementary material are still not very satisfactory, because SS3D~\citep{alwala2022pre} itself is struggling to create shapes with fine details. %Similarly, other results built upon DVR~\citep{niemeyer2020differentiable} with unsatisfactory quality are partial due to the same reason, see Figure~\ref{fig:wheels}. However, at the same time, our model can benefit from stronger SVR approaches in the future to achieve better performance. An example is shown in Figure~\ref{fig:im-net} where our model can generate shapes of higher quality built upon IM-Net~\citep{chen2019learning}. %Second, our text-guided stylization only works on texture generation but not on shapes, since we found that optimizing the implicit occupancy decoder $D_o$ with $L_{\text{C}}$ leads to degraded results. This problem needs further investigation in the future. 
Second, we cannot generate the categories outside the image dataset %In shape-and-texture stylization, since the initial shape derived from our two-stage feature-space alignment already provides a good 3D shape prior, we can produce reasonable 3D shapes even though we fine-tune the shape decoder in this process. However, this is not the case to generate novel categories beyond the image dataset 
due to the lack of 3D prior of the unseen category. That is why our model needs images as the stepping stone to learn what the particular category is like.  However, we want to highlight the following. First, built upon SS3D~\citep{alwala2022pre}, our approach can generate a wide range of categories with single-view images in the wild as training data. Second, with our shape-and-texture stylization, our approach can generate imaginary and uncommon shapes outside the image dataset (in the same category). Third, it is extremely challenging to generate arbitrary category shapes from text. As far as we know, there is only one existing work, Dream Field~\citep{jain2021zero}, that can generate more categories than ours. However, Dream Fields only generate multi-view images instead of directly generate 3D shapes, and it cannot generate reasonable shapes in many cases as shown in Figure 5 in our main paper and Figure~\ref{sec:ablation_supp} in this supplementary material.

\begin{comment}
\begin{figure*}
\centering
\includegraphics[width=0.99\textwidth]{wheels.pdf}
\vspace*{-1.5mm}
\caption{Our generative quality is limited by the DVR model that we build upon. (a) The result of DVR with single image as input. (b) Our result with text as input. }
\label{fig:wheels}
\vspace*{-2.5mm}
\end{figure*}
\end{comment}

\section{Text Set in the Experiments}~\label{sec:textset}

Recent works~\citep{chen2018text2shape,sanghi2021clip,jain2021zero} proposed their own text sets. However, their datasets have some limitations and are not suitable to evaluate our approach. The dataset of Text2shape~\citep{chen2018text2shape} contains text descriptions in only two categories, \ie, Table and Chair; CLIP-Forge~\citep{sanghi2021clip} lacks of descriptions on the color and texture; while Dream Fields~\citep{jain2021zero} utilizes text descriptions containing complex scenes and actions. To fairly evaluate our approach, we propose two text datasets on the ShapeNet~\citep{shapenet2015} and CO3D~\citep{reizenstein2021common} categories, respectively, shown in Tables~\ref{tab:shapenet} and~\ref{tab:co3d}.

\section{List of Notations}~\label{sec:notations}

In this section, we summarize symbols and notations used in the paper to facilitate readers to follow up. Please refer to Table~\ref{tab:notation} for more details.

\begin{table}
\centering
\caption{Summary of symbols used in paper. 
}
\label{tab:notation}
\scalebox{0.9}{
  \begin{tabular}{cc|cc}
    \toprule
      Notation & Description & Notation & Description  \\
      $E_S$ & encoder of SVR model & $E_I$ &  CLIP image encoder \\
      $E_T$ &CLIP text encoder   & $D$ & Decoder of SVR model \\
      $M$ & Mapper &  $M'$ & Mapper after stage-2 alignment \\
      $D_o$ &occupancy decoder& $D_c$ &color decoder \\
        $R$ & rendered images & $f_I$ & CLIP image feature \\
        $f_T$ & CLIP text feature & $f_S$ & shape feature in SVR model \\
      $o$ & camera center & $p$ & query point \\
      $d$ & cosine distance & $L_M$ &regression loss in stage 1\\
      $L_D$ & original loss in SVR model & $L_{bg}$& background loss\\
      $L_{bg\_1}$& background loss in stage 1 & $L_{bg\_2}$& background loss in stage 2\\
      $L_C$ & CLIP consistency loss & $L_{P}$& 3D prior loss\\
        $\Omega_T$ & CLIP text feature space & $\Omega_S$ & shape feature space from the SVR model \\
        $\Omega_I$ & CLIP image feature space\\
    \bottomrule
  \end{tabular}
  }
\end{table}

\begin{table}
\centering
\caption{Texts on ShapeNet~\citep{shapenet2015}. They are utilized to measure FID (Table 1 of the main paper), and employed in Human Perceptual Evaluation (Table~\ref{tab:consistent}) and A/B/C Test(Table~\ref{tab:ABC}). 
}
\label{tab:shapenet}
\scalebox{0.9}{
  \begin{tabular}{cc}
    \toprule
      a glass single leg circular table &
a wooden double layers table \\
a square metal table&
a round shaped single legged wooden table \\
this is a bar stool with metal arches as a design feature&
a children chair with little legs\\
a swivel chair with wheels&
a red recliner seems comfortable\\
a red car&
a green SUV\\
a large black truck&
a long luxury black car\\
army fighter jet&
a black airplane with long white wings\\
a blue airplane with short wings&
boeing 747\\
a big ship for transportation&
a boat with sail\\
a watercraft&
a wooden boat\\
a blue sofa&
sofa with legs\\
a sofa with black backrest&
a small sofa\\
a long brown bench&
a marble bench\\
a metal bench&
concrete bench\\
a military sniper rifle with long barrel&
a rifle with magazines\\
a short rifle&
rifle shotgun\\
a computer monitor&
a display\\
a monitor with square base&
a TV monitor\\
a cabinet with cylindrical legs&
a cupboard\\
a long brown cabinet&
a wardrobe\\
a desk lamp&
bedside lamp\\
lamp supported by a long pillar&
mushroom-like lamp\\
a mobile phone&
a small cell phone\\
a mobile phone with black screen&
an iphone\\
a columnar loudspeaker&
a loudspeaker with metal surface\\
a wooden loudspeaker&
a cylindrical loudspeaker\\
    \bottomrule
  \end{tabular}
  }
\end{table}

\begin{table}
\centering
\caption{Texts on CO3D~\citep{reizenstein2021common}.}
\label{tab:co3d}
\scalebox{0.74}{
  \begin{tabular}{cccc}
    \toprule
    A big apple & A red apple &
    A green bottle  & A tall cylindrical bottle \\
    A white cup & A wooden cup &
    A large black microwave & A white cuboid microwave \\
    A black skateboard & A green long skateboard &
    A cute toytruck & A large toy truck \\
    A blue backpack & A red big backpack &
    A white bowl& A big wooden bowl \\
    A red round frisbee & A blue large frisbee&
    A big blue motorcycle & A black large wheels motorcycle\\
    A circular stop sign & A triangle stop sign &
    Tv screen & A grey big tv screen \\
    A basketball & A tennis ball &
    A large broccoli  & A green broccoli \\
    A hairdryer & A yellow hairdryer&
    A black mouse & A white mouse\\
    A cuboid big suitcase & A large size tall suitcase &
    A round umbrella  & A big black umbrella\\
    A big banana & A long banana&
    A cream round cake & A chocolate mooncake \\
    A blue handbag & A red big handbag&
    An orange  & A large round orange \\
    A teddybear & A cute teddybear &
    A blue fat vase & A blue tall vase \\
    A black baseball bat & A long wooden baseball bat &
    A blue car & A red car\\
    An egg hotdog & A sausage hotdog&
    A black parkingmeter & A white tall parkingmeter \\
    A black toaster & A round toaster&
    Tall wineglass & Single leg big wineglass\\
    A brown baseball glove & A black big baseball glove &
    A big carrot & A long carrot\\
    A red hydrant & A yellow hydrant&
    A large round pizza & A tomato meat pizza\\
    A white toilet & A fat white toilet&
    A stone bench & A wooden long bench \\
    A gray iphone & A black phone &
    A long black keyboard & A short white keyboard\\
    A short tree & A tall green tree &
    A toy bus & One decker toy bus \\
    A blue bicycle & A black bicycle &
    A blue chair & A wooden chair \\
    A red kite & A long blue kite &
    A TV remote & A long white remote \\
    A book with blue cover & A black book&
    Brown couch & A long brown couch\\
    A open laptop & A black laptop&
    An egg sandwich & A meat sandwich \\
    A cute toy train & A short blue toy train &
    Chocolate donut & Big circular donut \\
    \bottomrule
  \end{tabular}
  }
\end{table}  

\appendix

\bibliography{iclr2023_conference}
\bibliographystyle{iclr2023_conference}
\end{document}